\documentclass[letterpaper, 10 pt, journal, twoside]{IEEEtran}

\IEEEoverridecommandlockouts                              

\usepackage{amsmath} 
\usepackage{amssymb} 

\usepackage{float}
\usepackage{graphicx}
\usepackage{amsmath}

\newcommand{\vect}[1]{\boldsymbol{#1}}
\newcommand{\mat}[1]{\boldsymbol{#1}}

\newcommand{\diffs}[3]{\frac{\partial^2 #1}{
\ifx#2#3 
\partial #2^2
\else
\partial #2 \partial #3
\fi
}}

\newcommand{\zerov}{\vect{0}}

\newcommand{\fv}{\vect{f}}
\newcommand{\gv}{\vect{g}}

\newcommand{\hv}{\vect{h}}
\newcommand{\kv}{\vect{k}}

\newcommand{\nv}{\vect{n}}

\newcommand{\qv}{{\vect{q}}}
\newcommand{\dqv}{\dot{\vect{q}}}
\newcommand{\ddqv}{\ddot{\vect{q}}}

\newcommand{\uv}{\vect{u}}

\newcommand{\xv}{\vect{x}}

\newcommand{\epsilonv}{\vect{\varepsilon}}

\newcommand{\etav}{\vect{\eta}}
\newcommand{\deltav}{\vect{\delta}}

\newcommand{\tauv}{\vect{\tau}}

\newcommand{\omegav}{\vect{\omega}}

\newcommand{\Mv}{\vect{M}}

\newcommand{\Xv}{\vect{X}}
\newcommand{\Yv}{\vect{Y}}

\newcommand{\Bm}{\mat{B}}

\newcommand{\Km}{\mat{K}}
\newcommand{\Mm}{\mat{M}}

\newcommand{\Qm}{\mat{Q}}
\newcommand{\Rm}{\mat{R}}

\newcommand{\Xm}{\mat{X}}
\newcommand{\Ym}{\mat{Y}}

\newcommand{\Sigmam}{\mat{\Sigma}}

\usepackage[export]{adjustbox}
\usepackage{xcolor}
\usepackage[normalem]{ulem}
\usepackage{booktabs}

\usepackage{balance} 

\def\Real{\mathbb{R}}

\def\SwingUpPend{\centering
    \includegraphics[width=0.116\textwidth]{Images/a_cut.png}
    \includegraphics[width=0.116\textwidth]{Images/exp1_b_cut.png} 
    \includegraphics[width=0.116\textwidth]{Images/exp1_c_cut.png}
    \includegraphics[width=0.116\textwidth]{Images/exp1_d_cut.png}
    \vskip 0.4em
    \includegraphics[width=0.116\textwidth]{Images/a_cut.png}
    \includegraphics[width=0.116\textwidth]{Images/exp2_b_cut.png} 
    \includegraphics[width=0.116\textwidth]{Images/exp2_c_cut.png}
    \includegraphics[width=0.116\textwidth]{Images/exp2_d_cut.png}}

\def\BlockDiagram{\centerline{\includegraphics[width=1\textwidth]{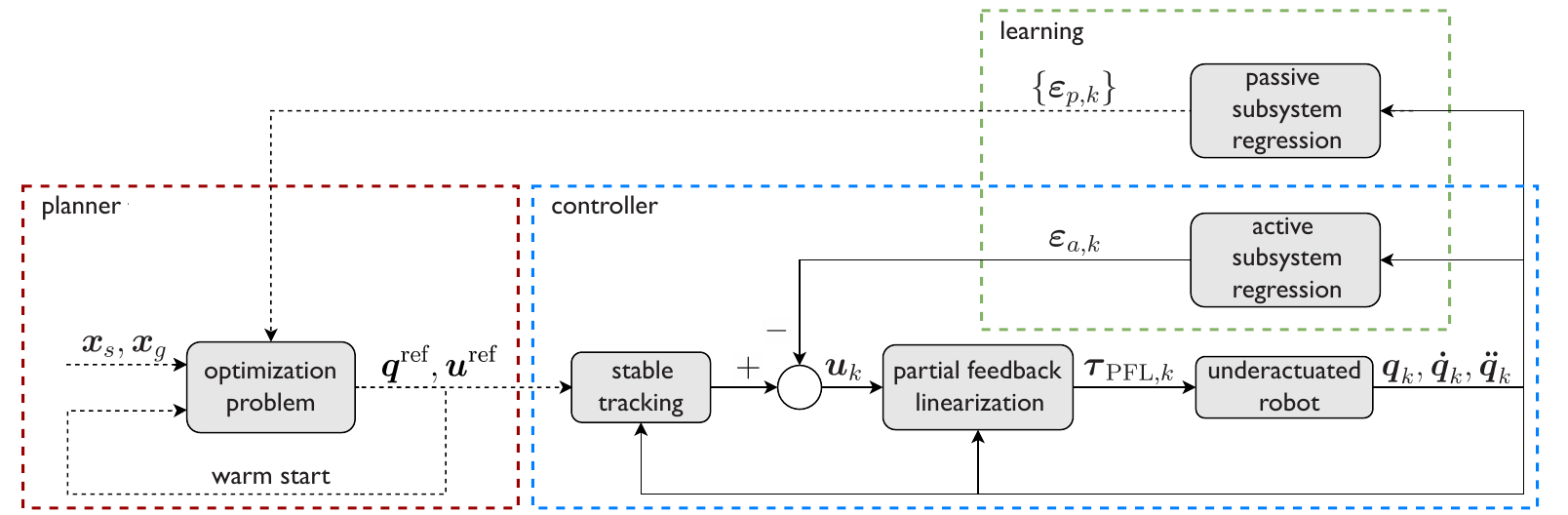}}}

\def\Pendubot{\centerline{\includegraphics[width=0.35\textwidth]{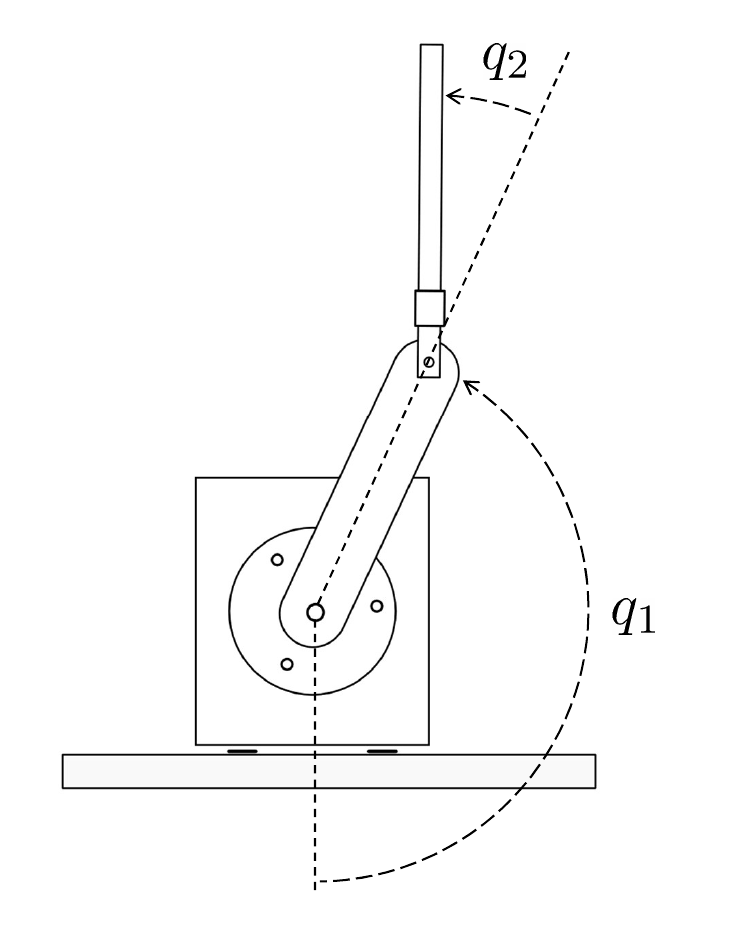}}}

\def\SwingUpSim{ \centering{
\includegraphics[width=0.235\textwidth]{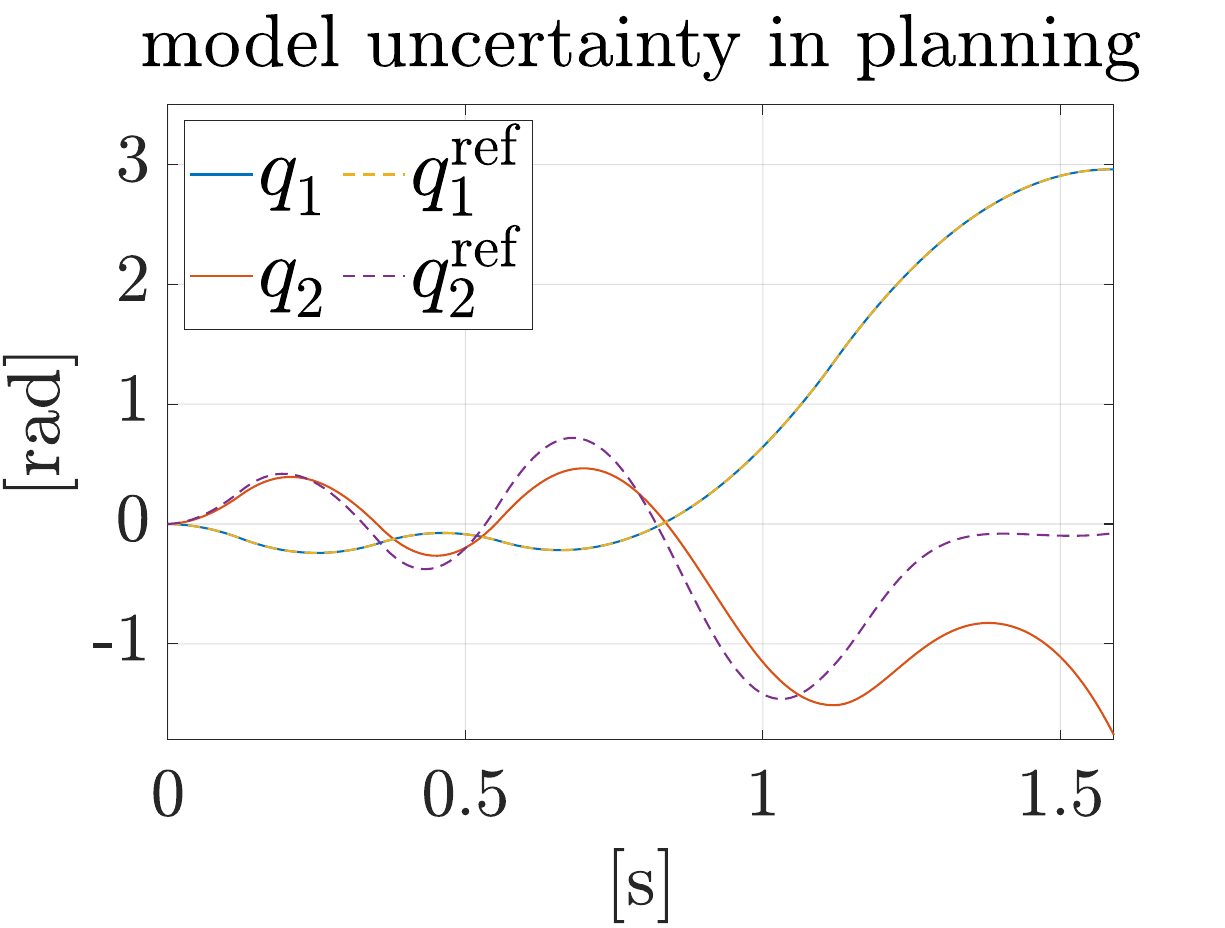}
\includegraphics[width=0.235\textwidth]{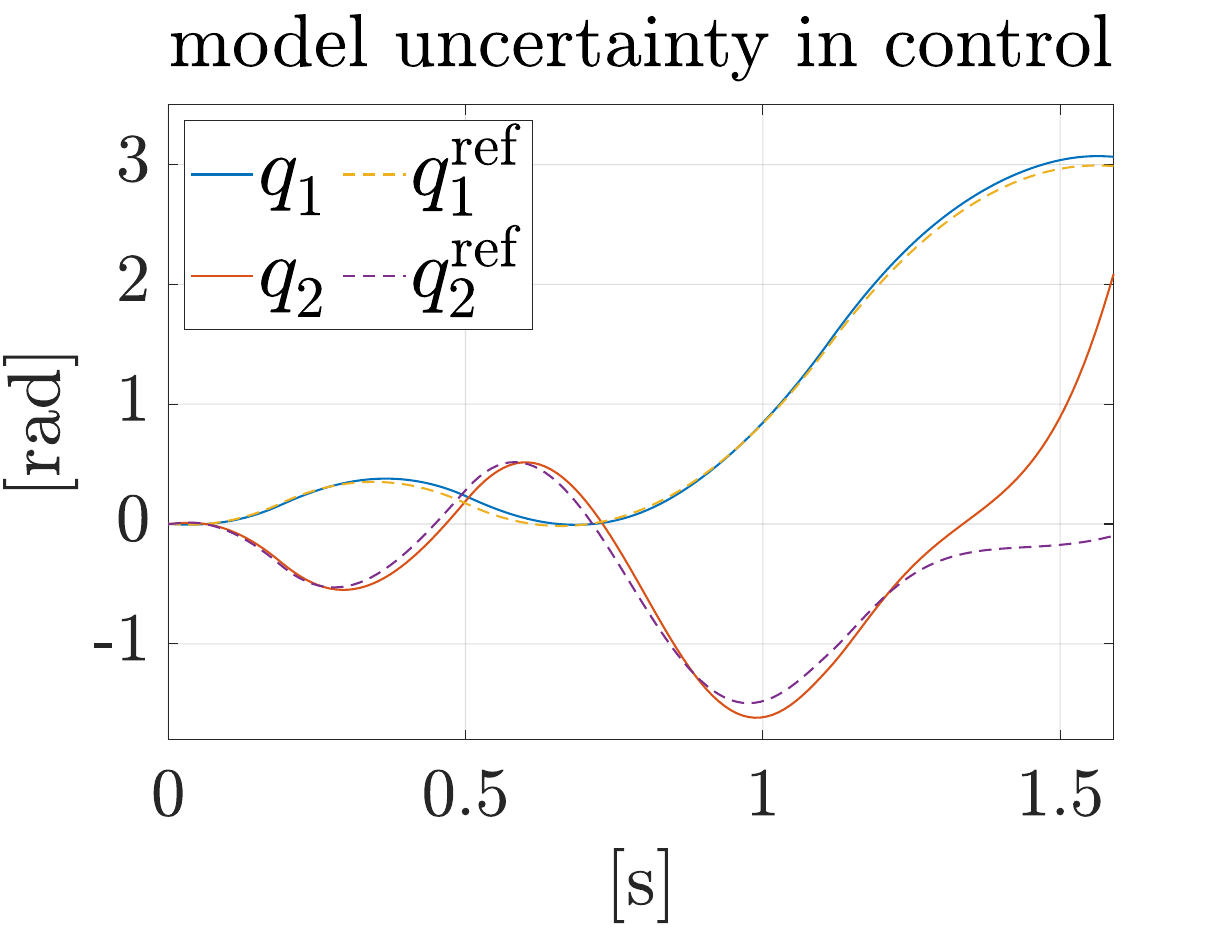}}}

\def\IterationsSim{\centering{
\includegraphics[width=0.235\textwidth]{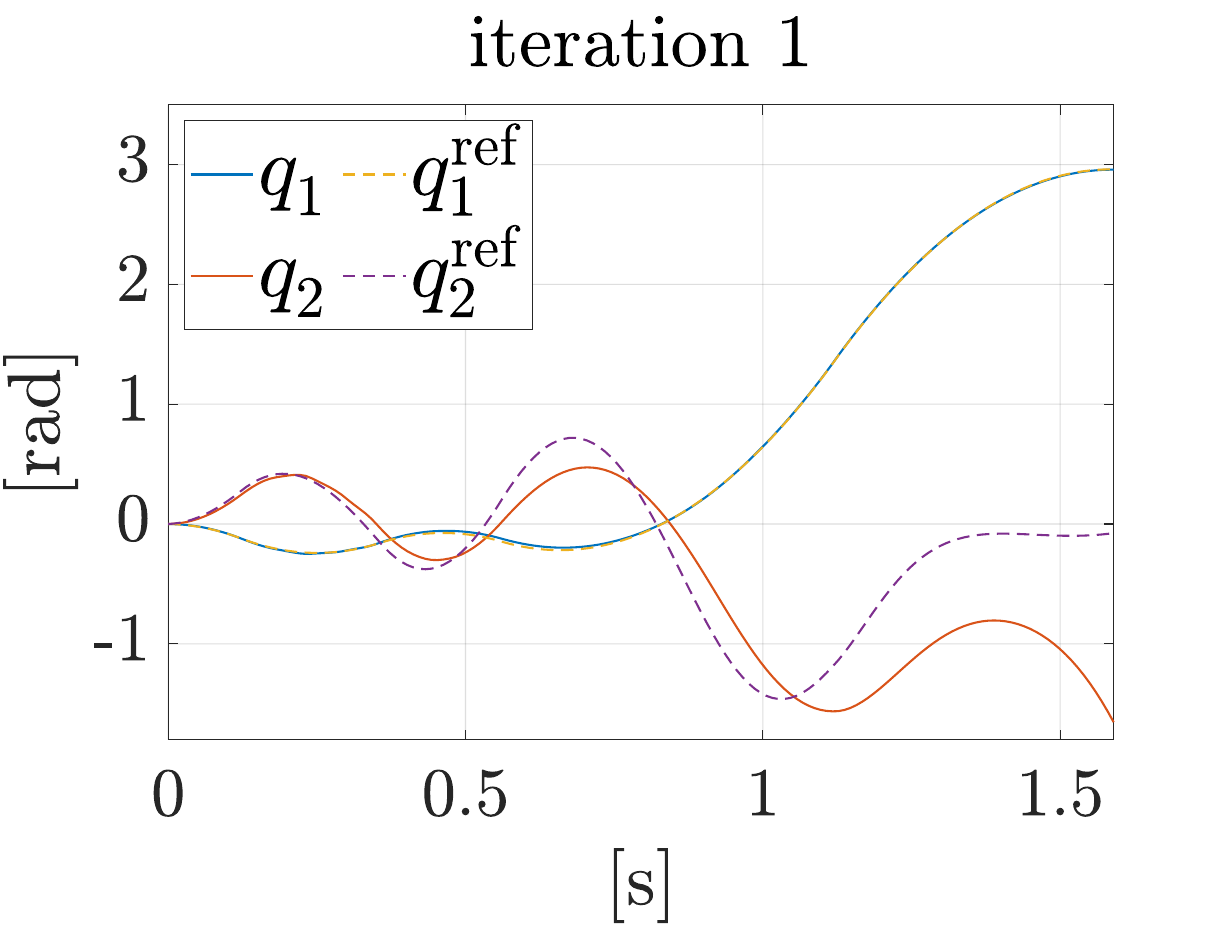}
\includegraphics[width=0.235\textwidth]{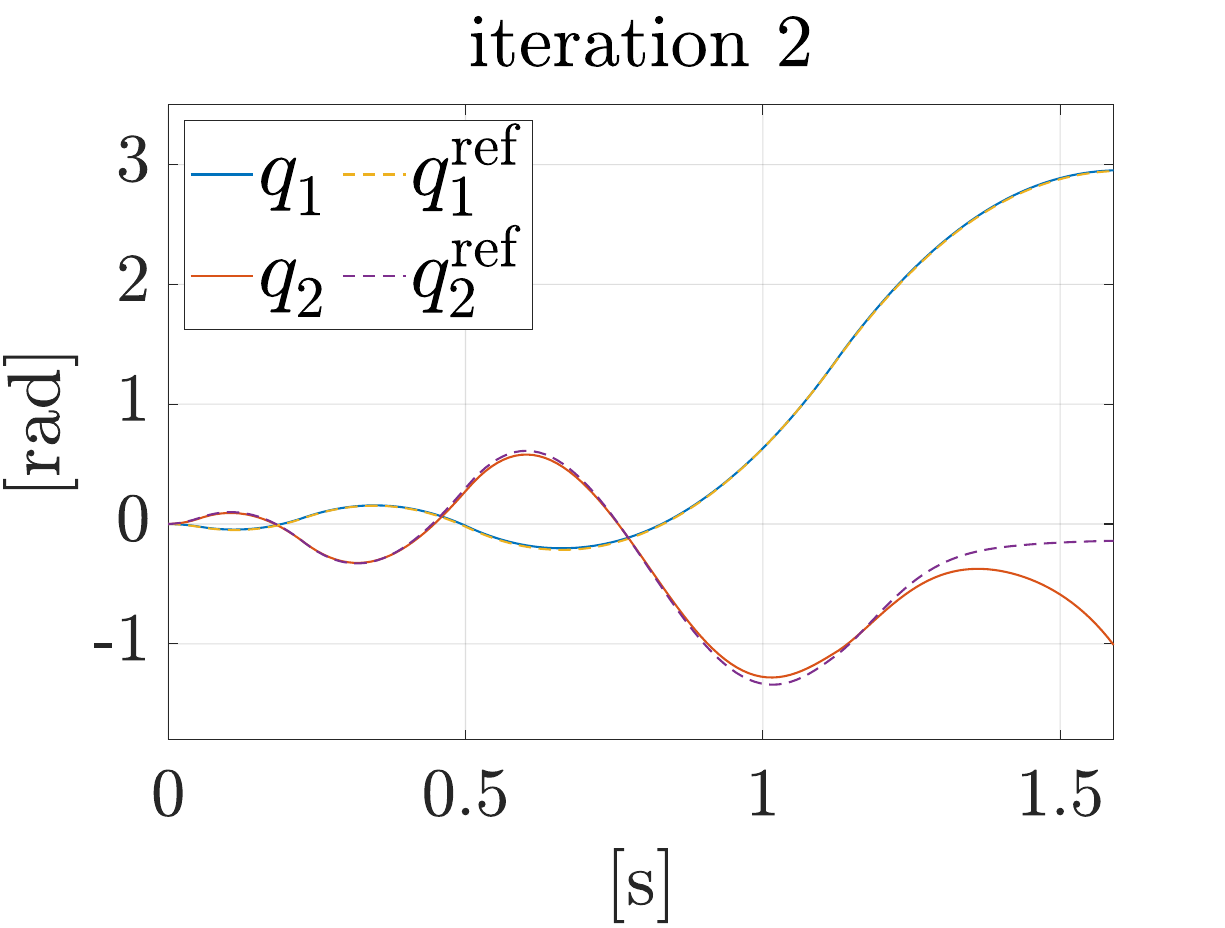}
\includegraphics[width=0.33\textwidth]{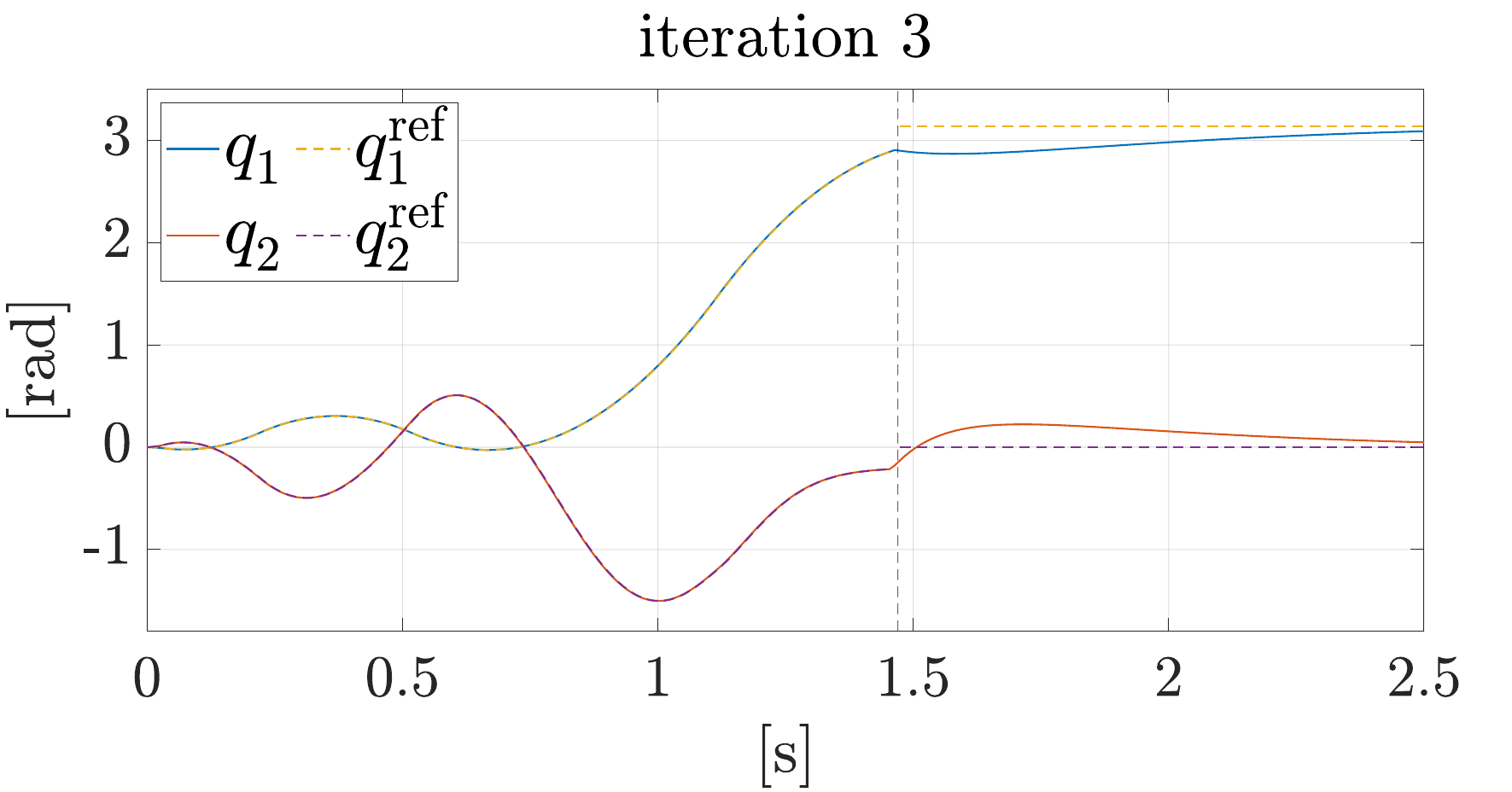}}}

\def\ShiriaevSim{\centering{
\includegraphics[width=0.235\textwidth]{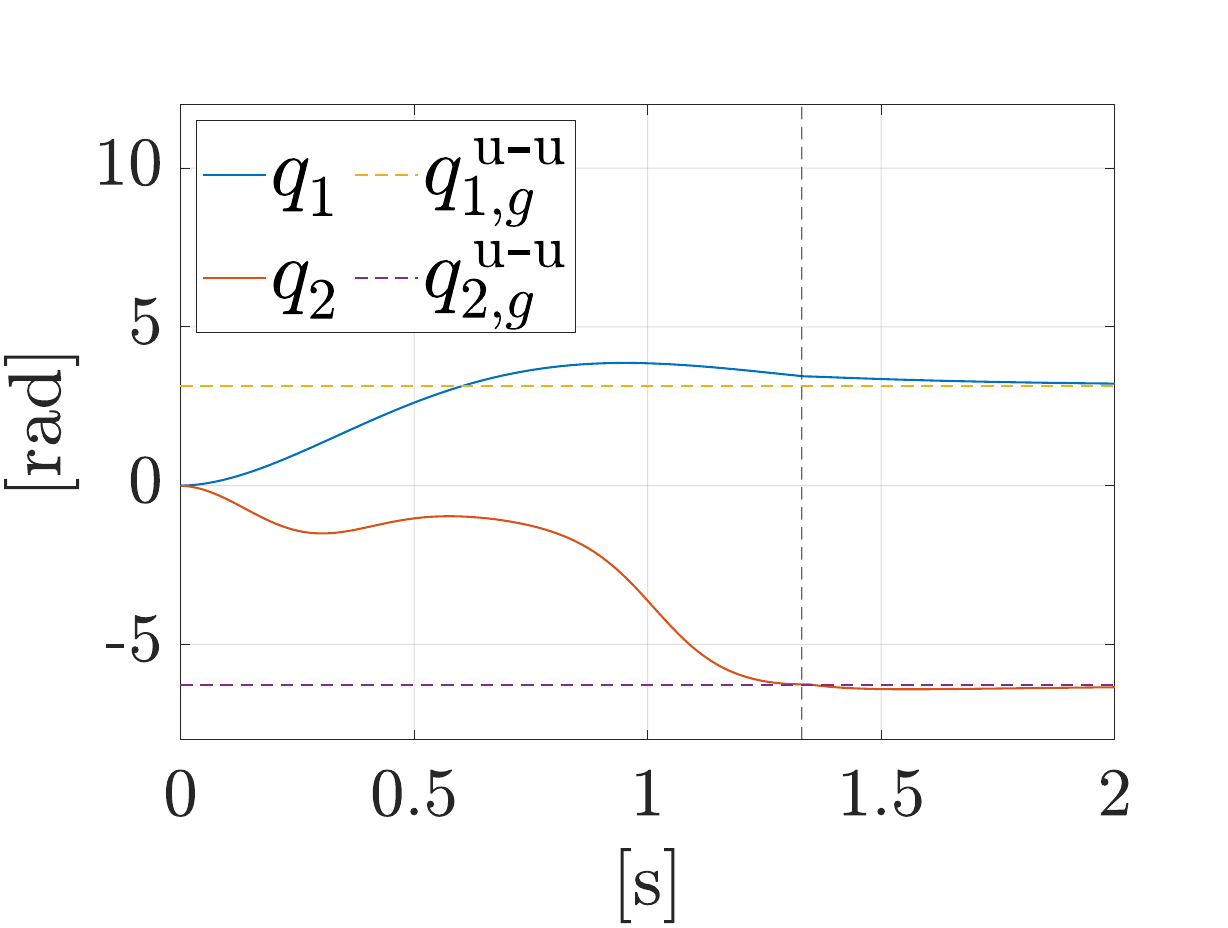}
\includegraphics[width=0.235\textwidth]{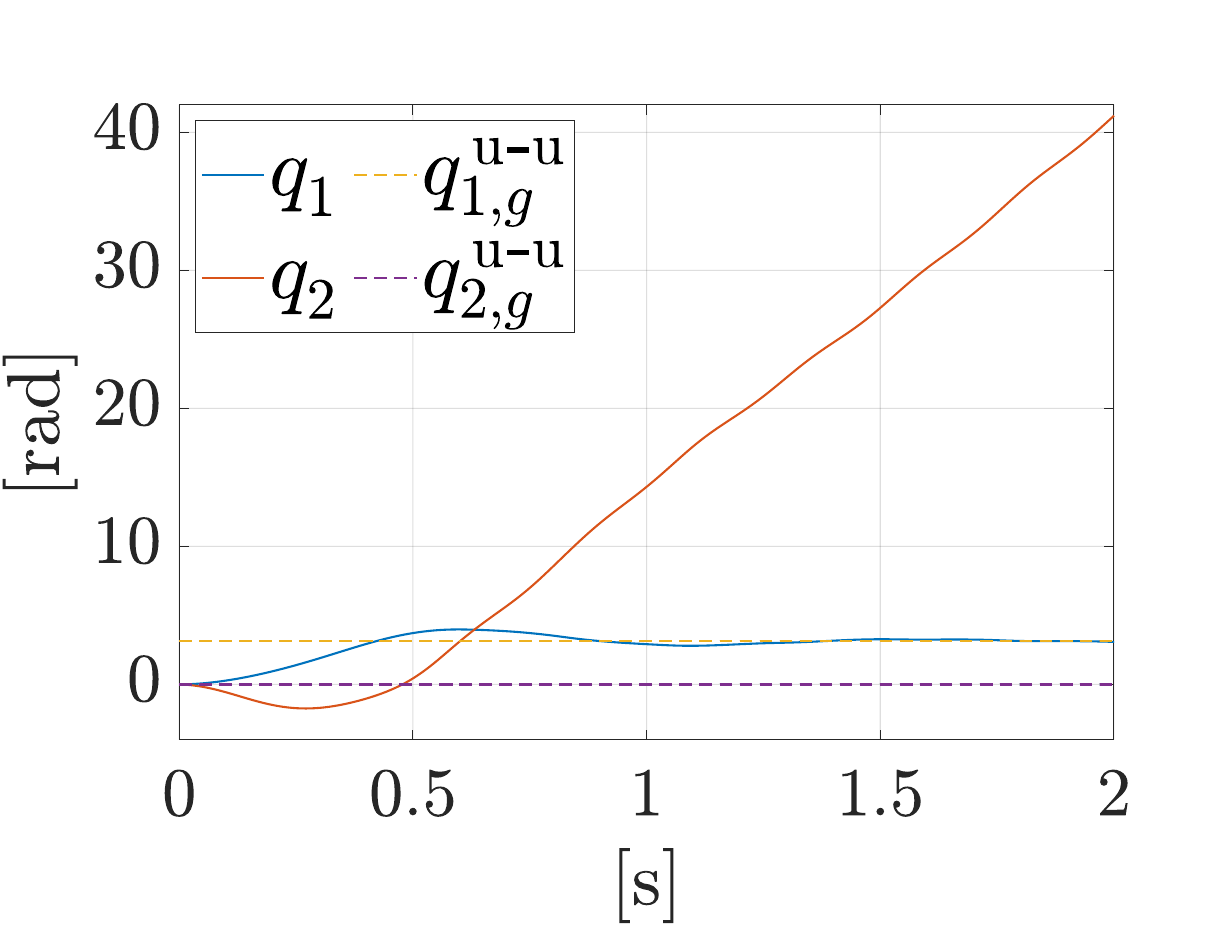}}}

\def\IterationsSimForcedEq{\centering{
\includegraphics[width=0.21\textwidth]{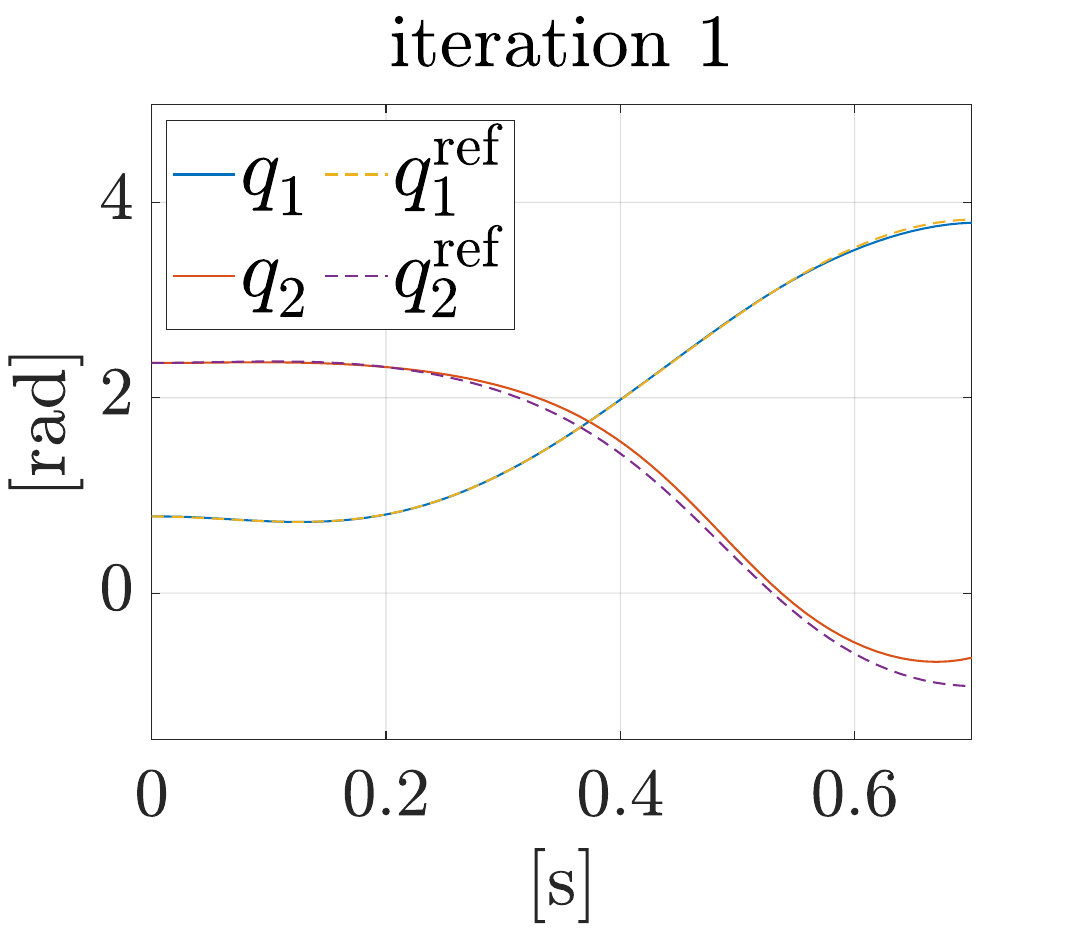}
\includegraphics[width=0.264\textwidth]{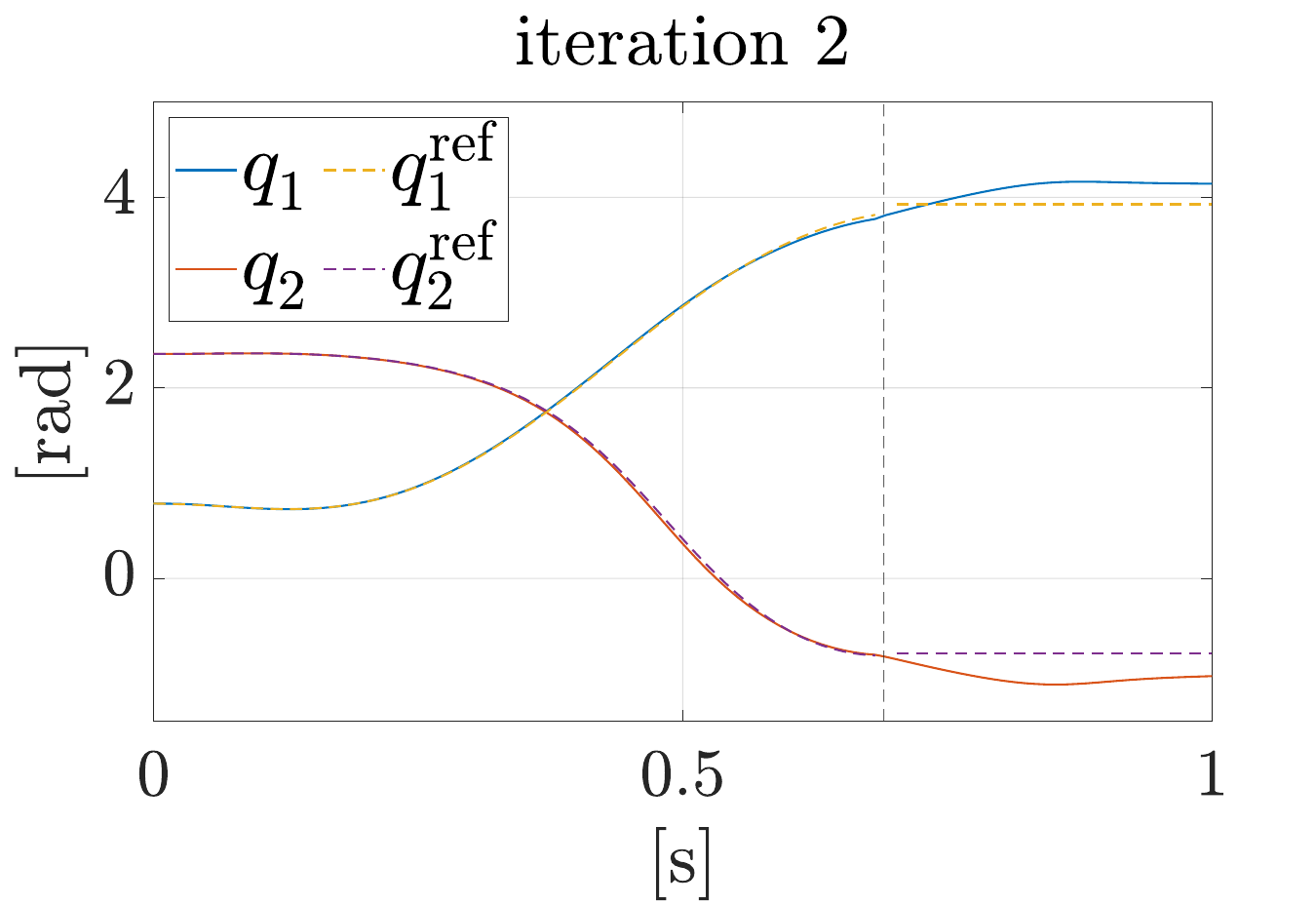}}}

\def\ExperimentUpUp{\centering{
\includegraphics[width=0.235\textwidth]{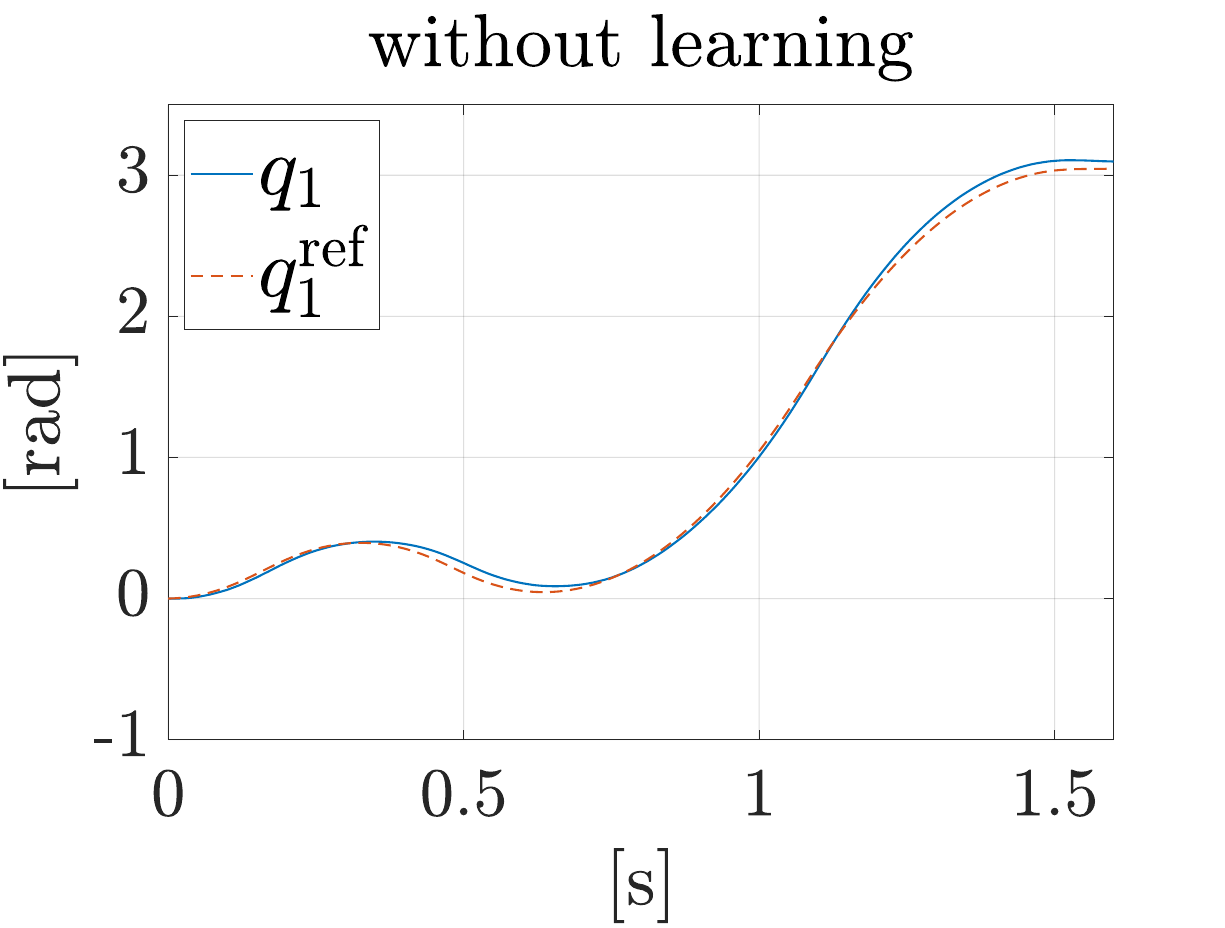}
\includegraphics[width=0.235\textwidth]{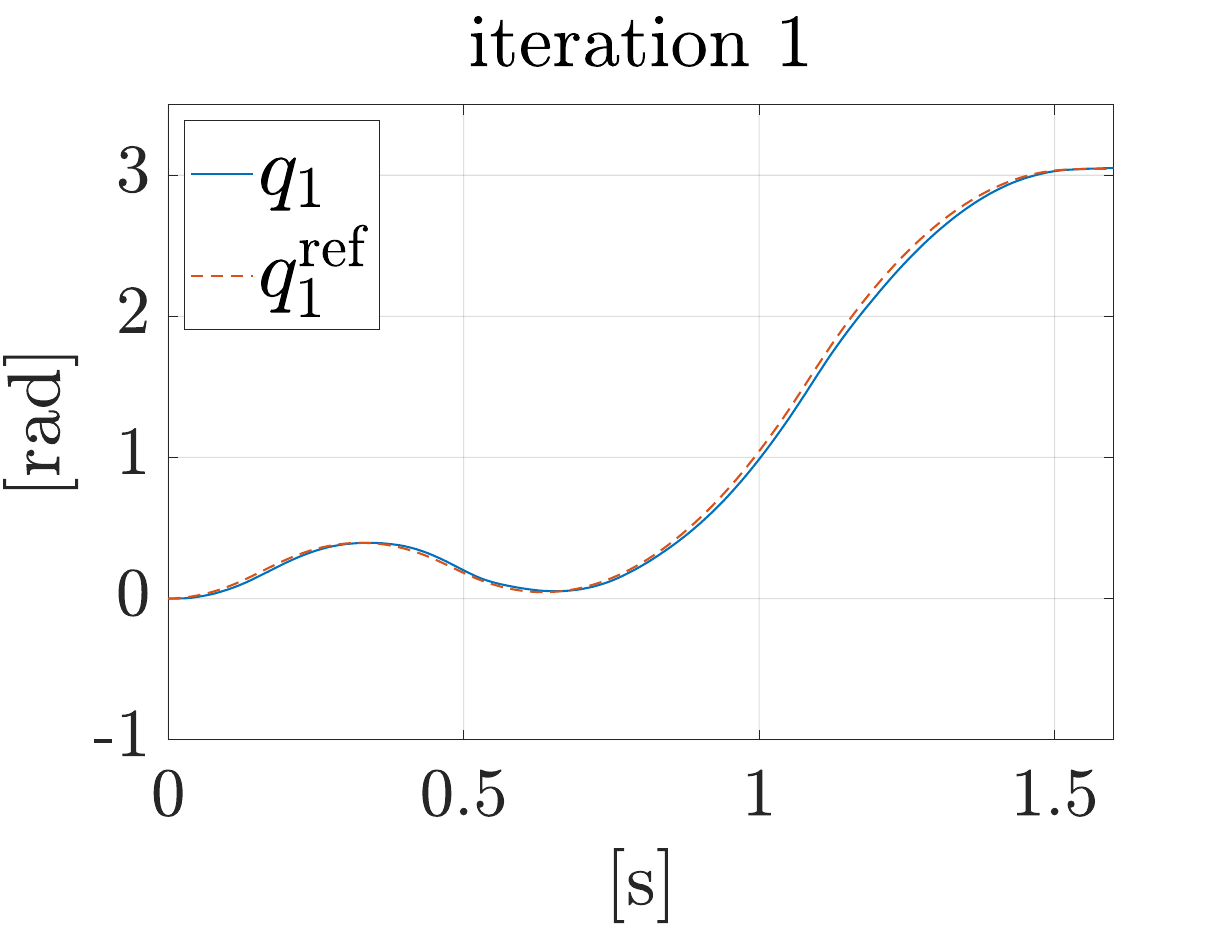}
\includegraphics[width=0.235\textwidth]{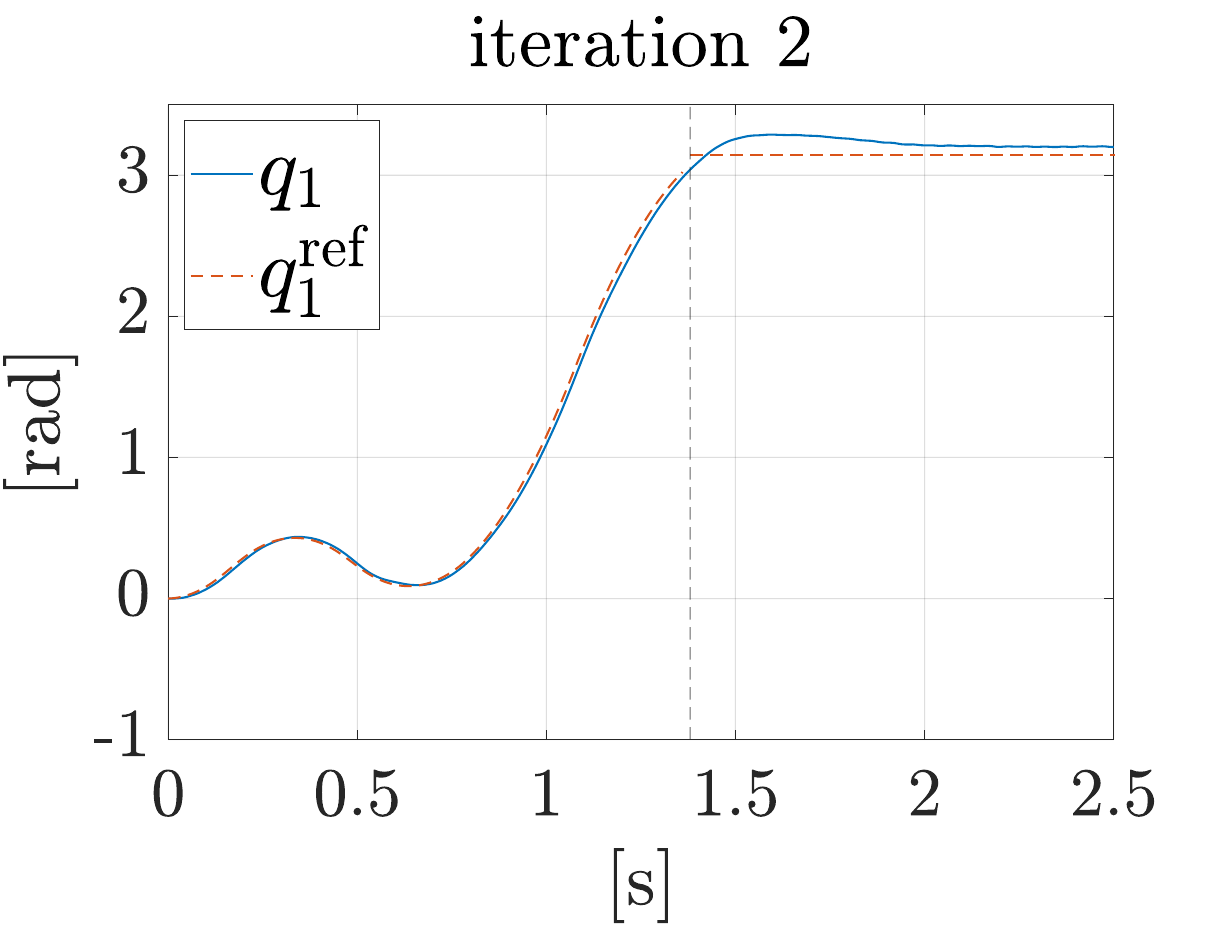}
\includegraphics[width=0.235\textwidth]{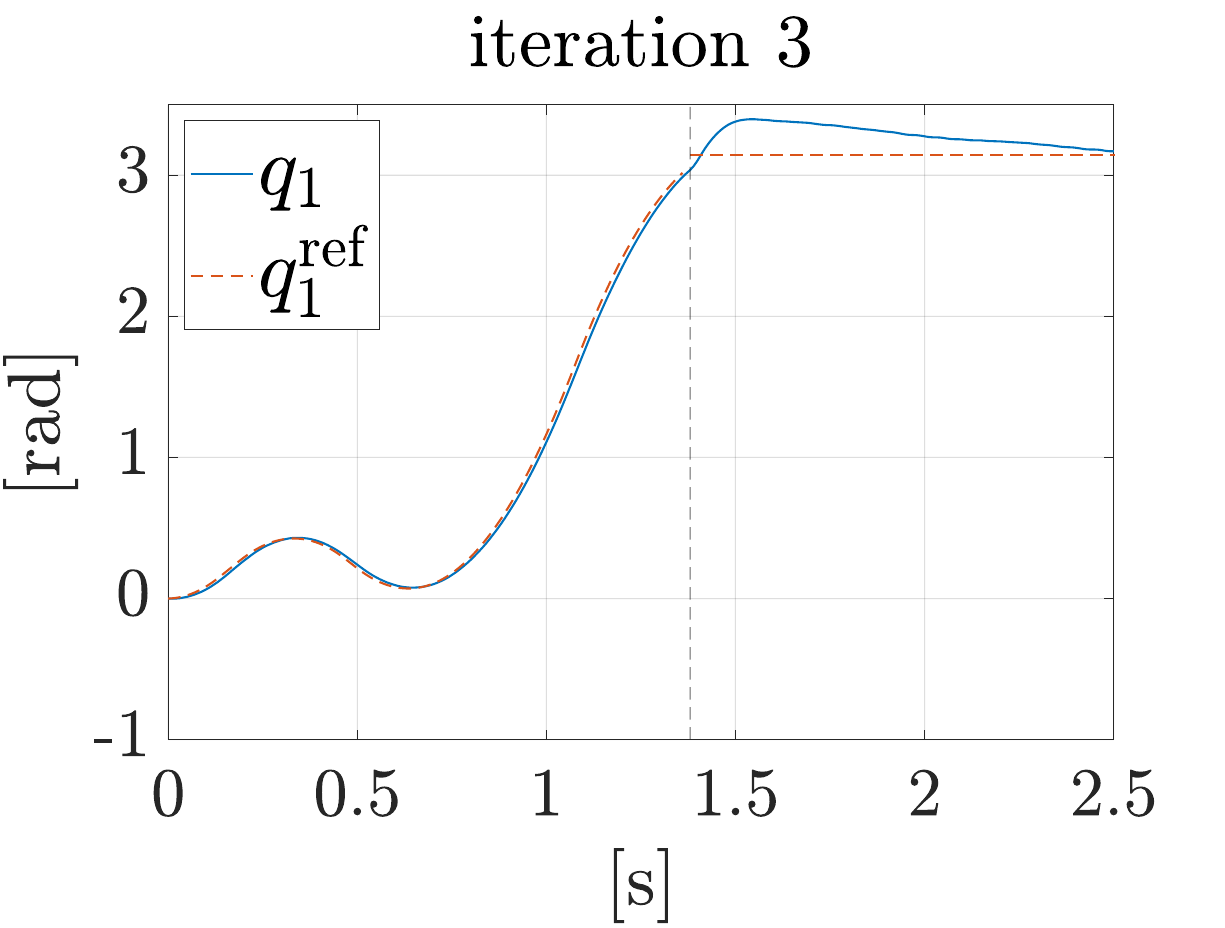}
}
\centering{
\includegraphics[width=0.235\textwidth]{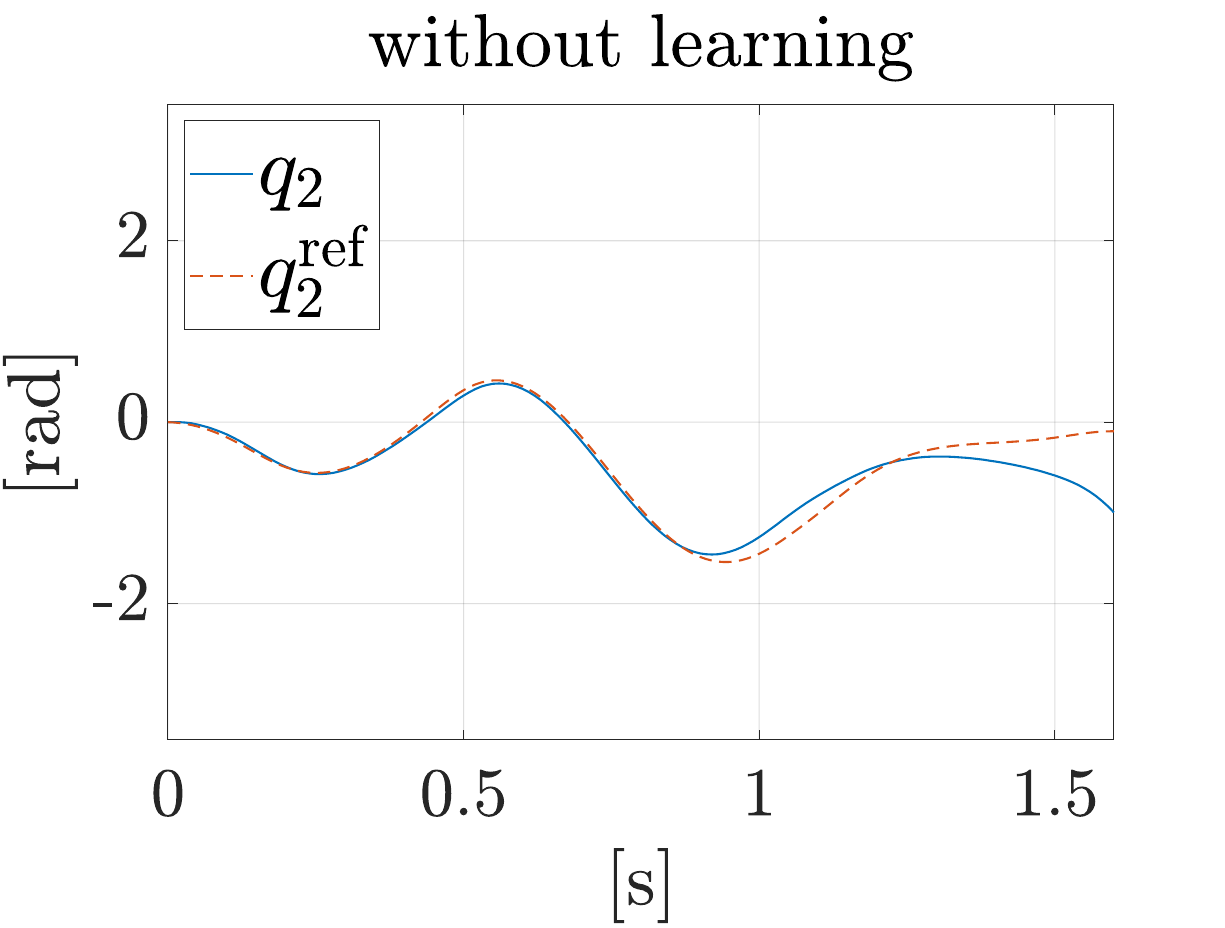}
\includegraphics[width=0.235\textwidth]{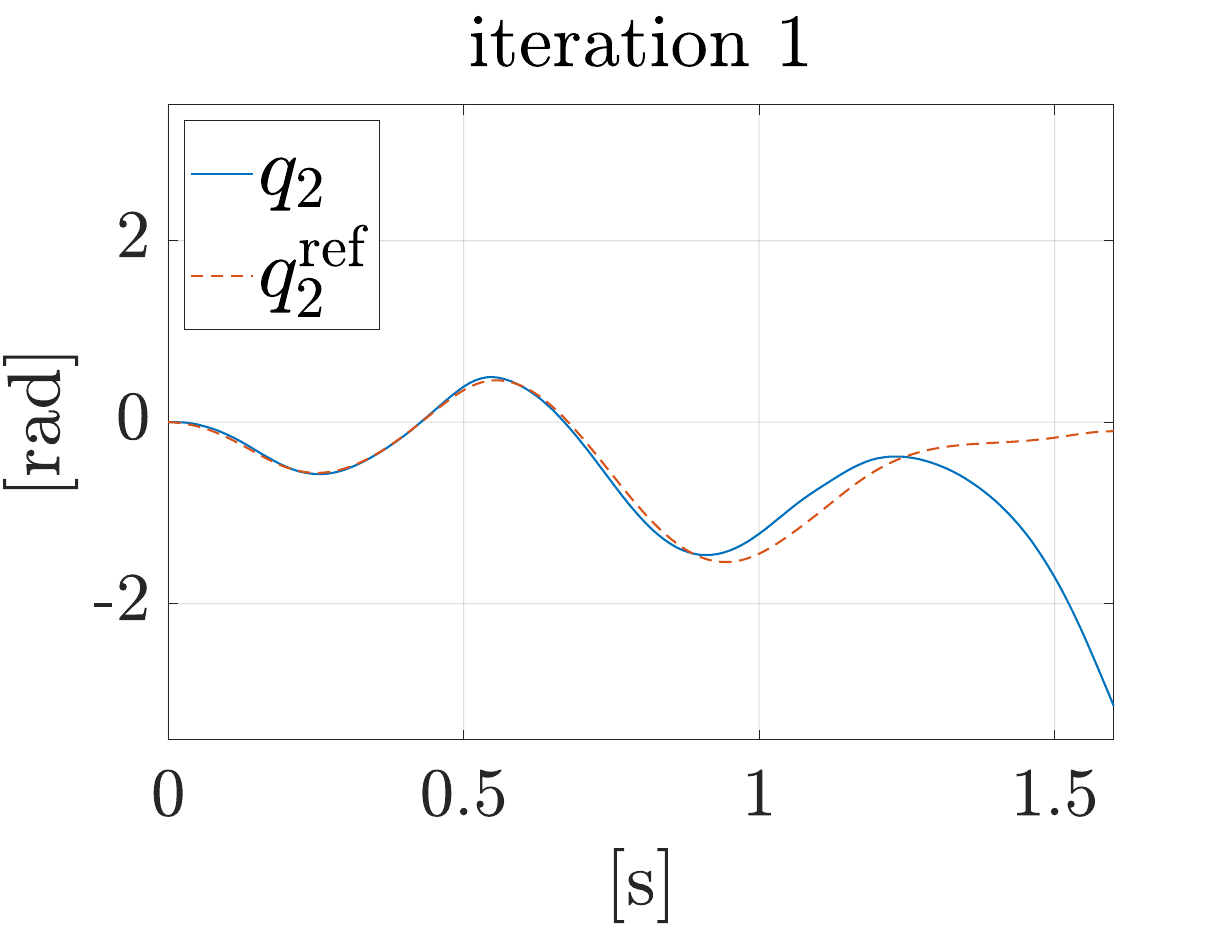}
\includegraphics[width=0.235\textwidth]{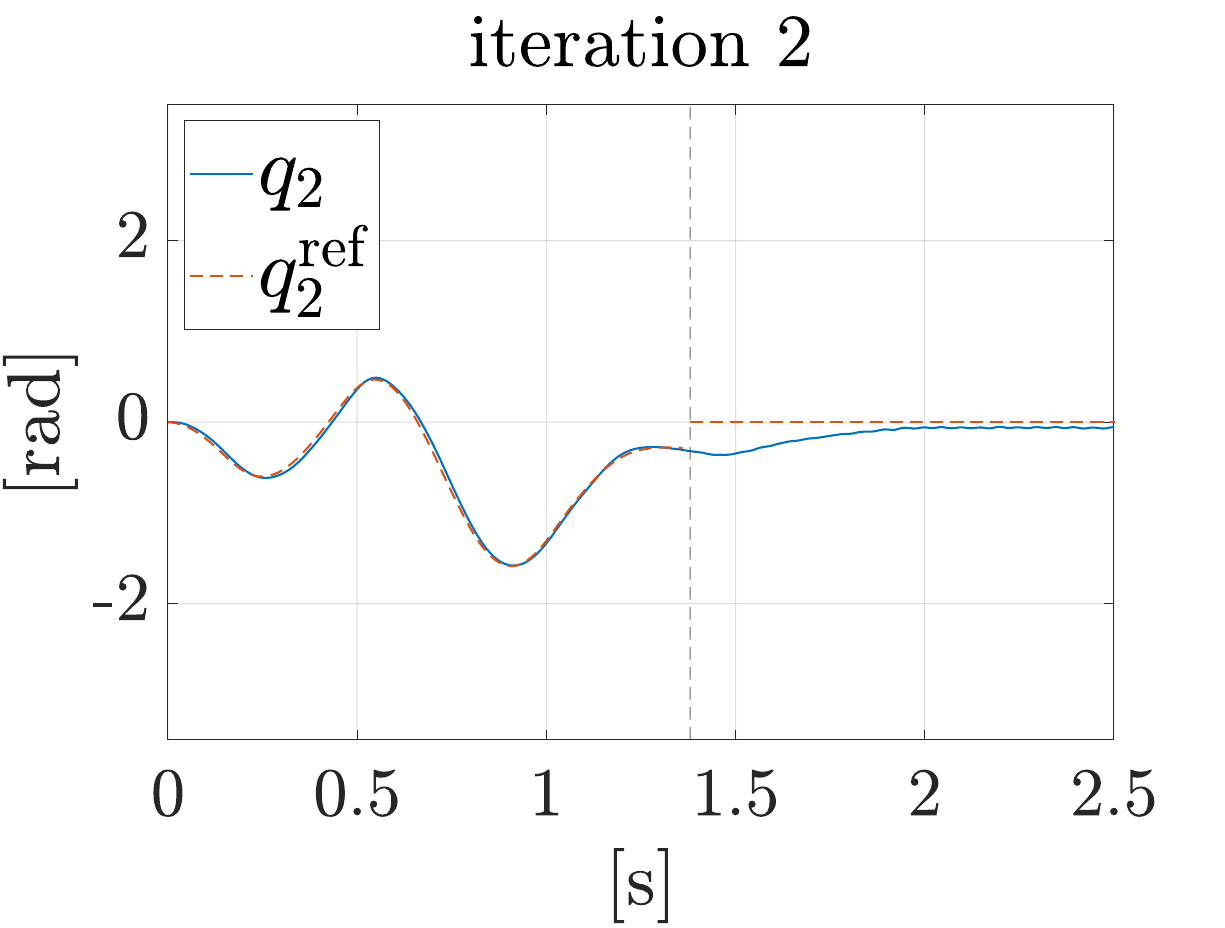}
\includegraphics[width=0.235\textwidth]{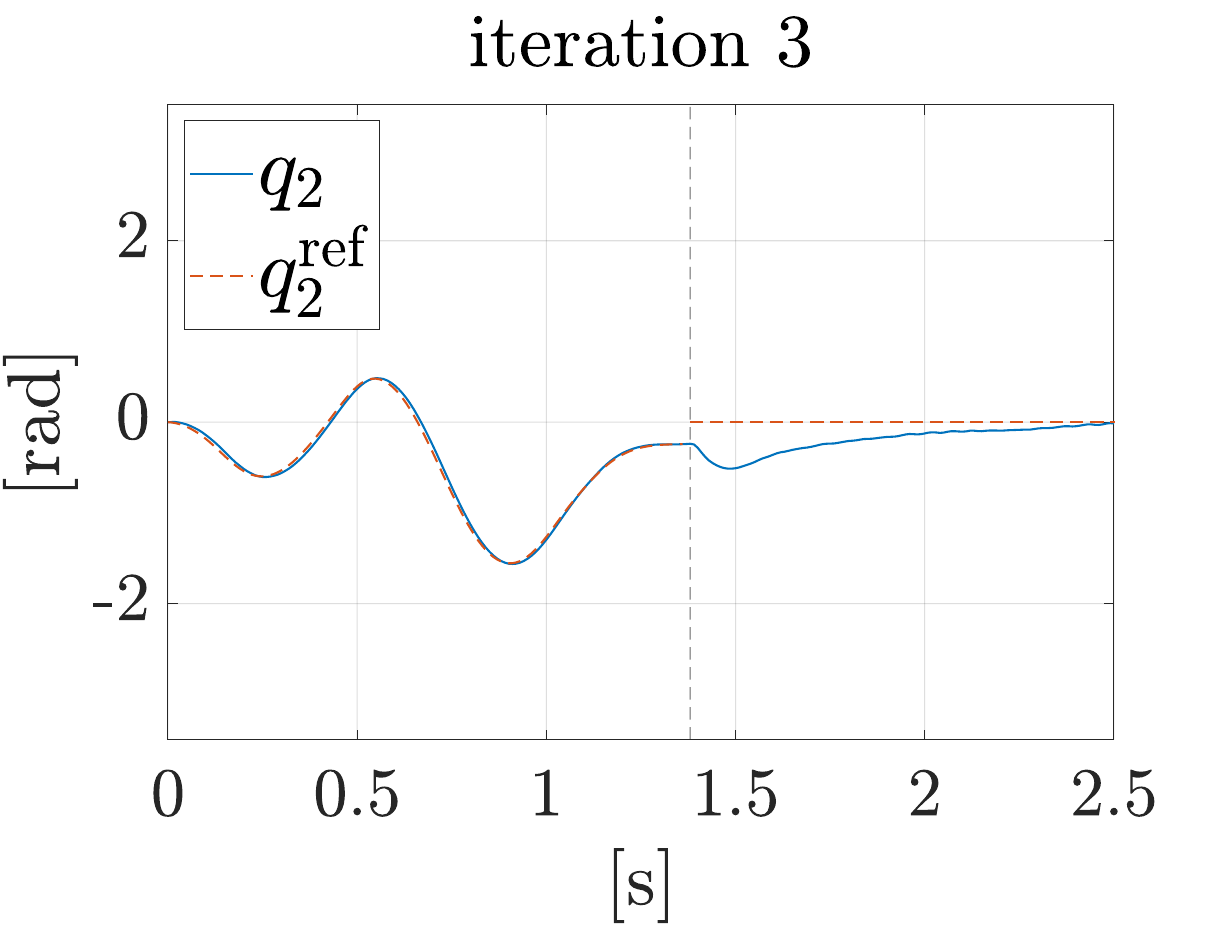}
}}

\def\ShiriaevExperiment{\centering{
\includegraphics[width=0.235\textwidth]{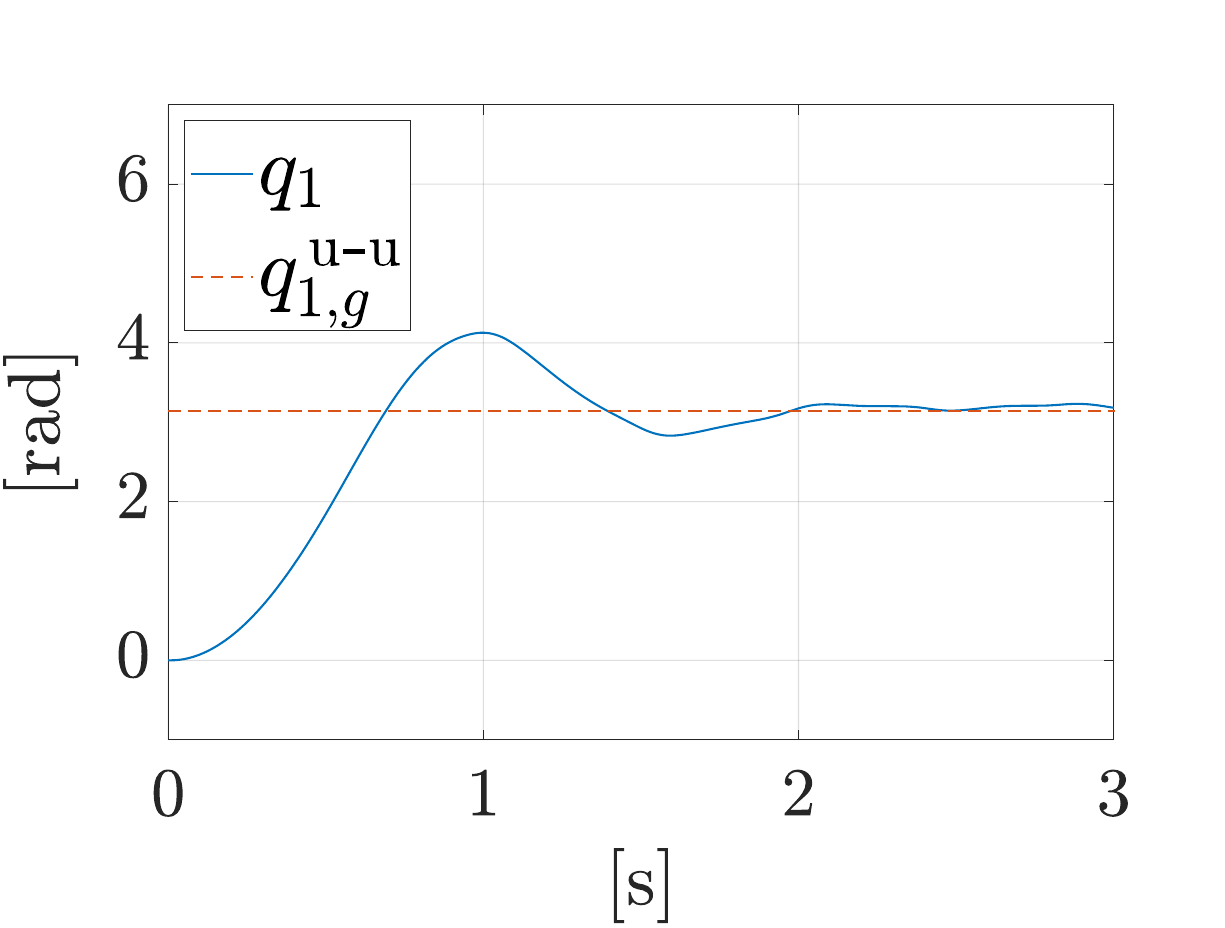}
\includegraphics[width=0.235\textwidth]{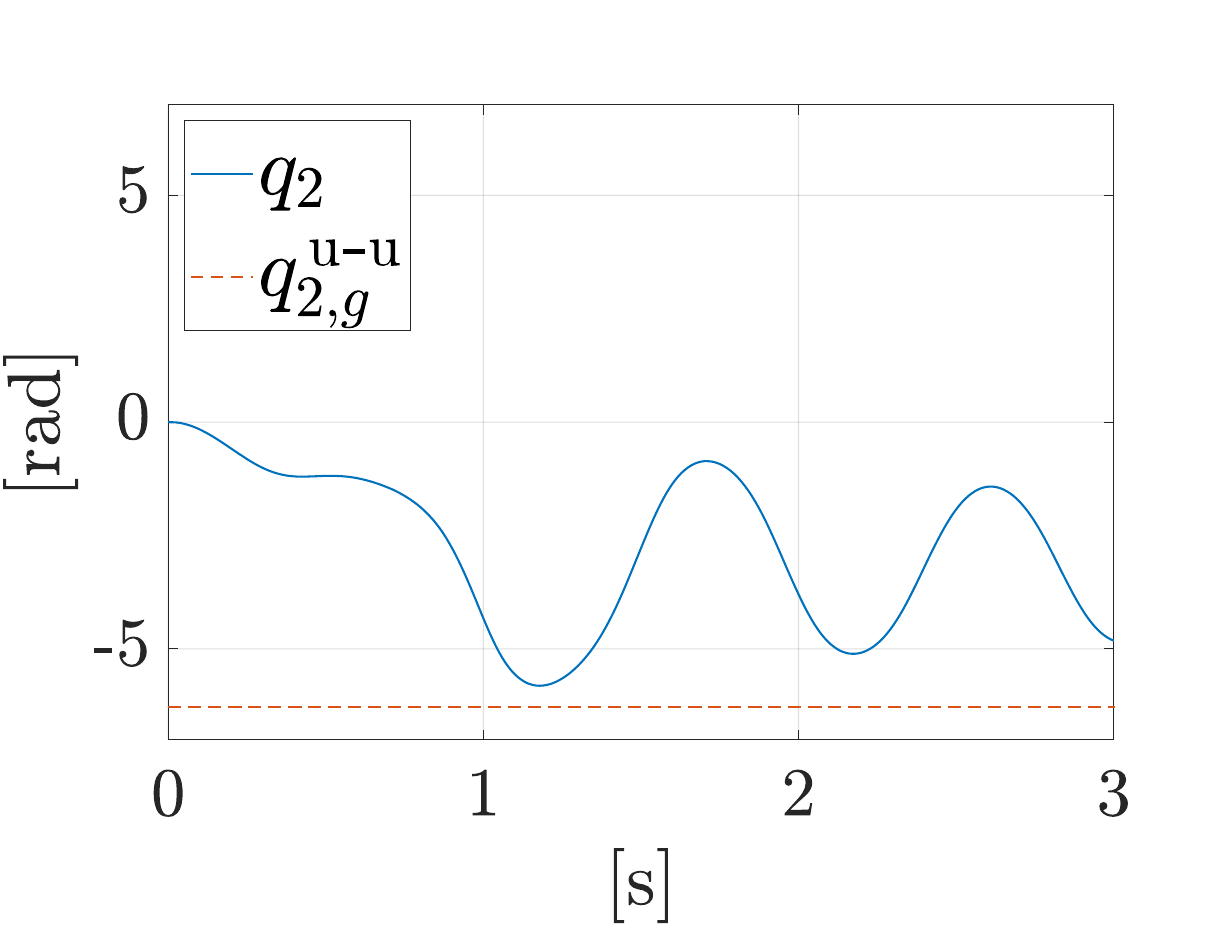}
}}

\def\ExperimentDownUp{\centering{
\includegraphics[width=0.235\textwidth]{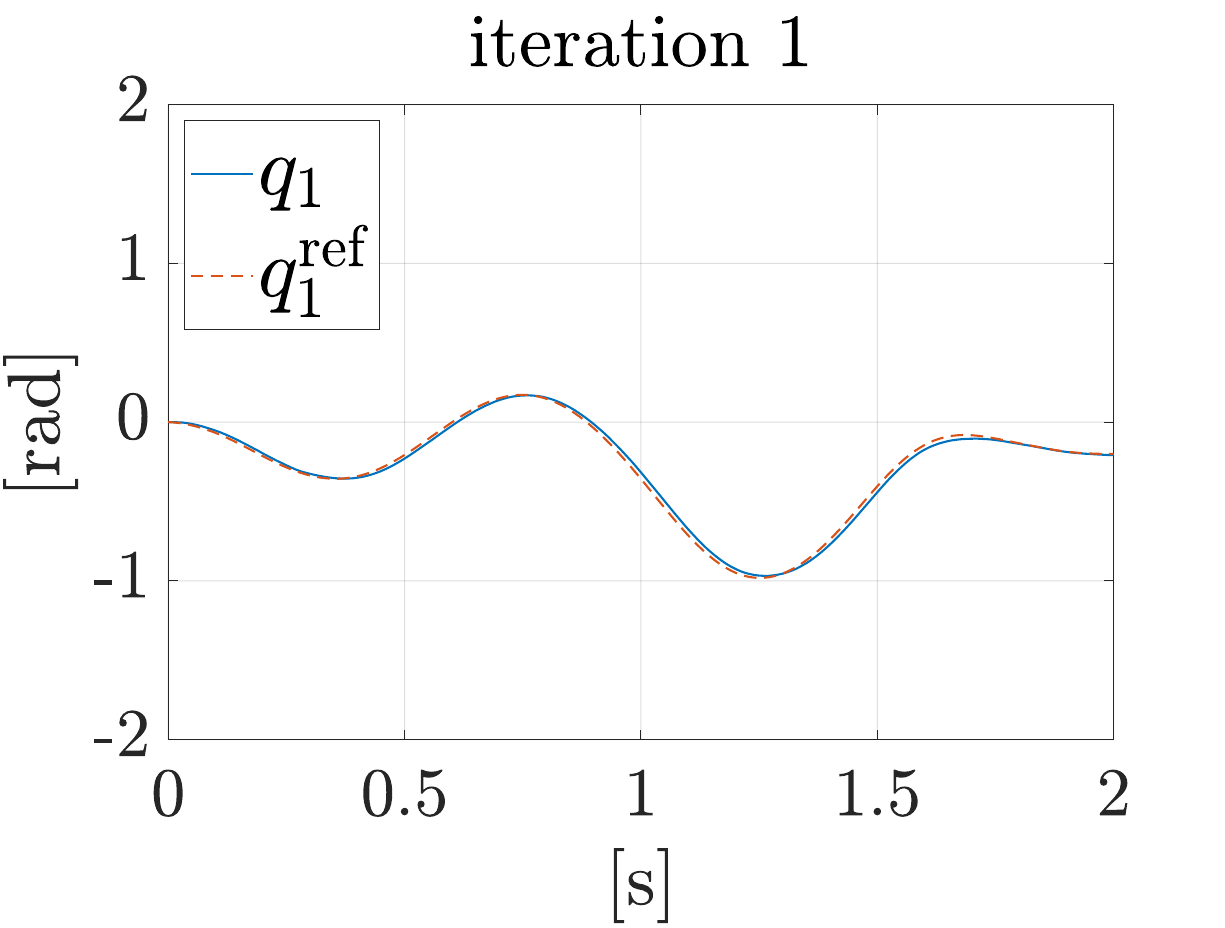}
\includegraphics[width=0.235\textwidth]{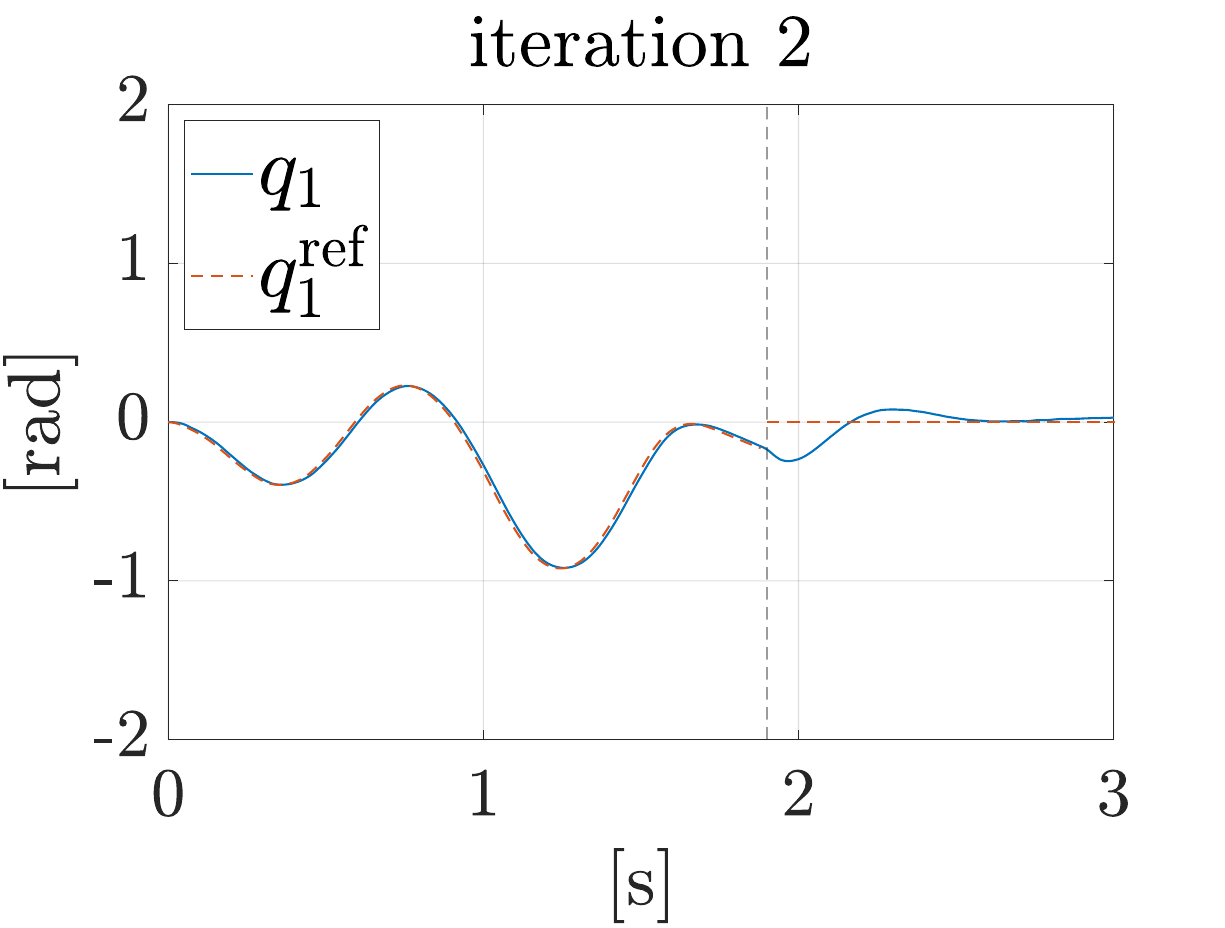}}
\centering{
\includegraphics[width=0.235\textwidth]{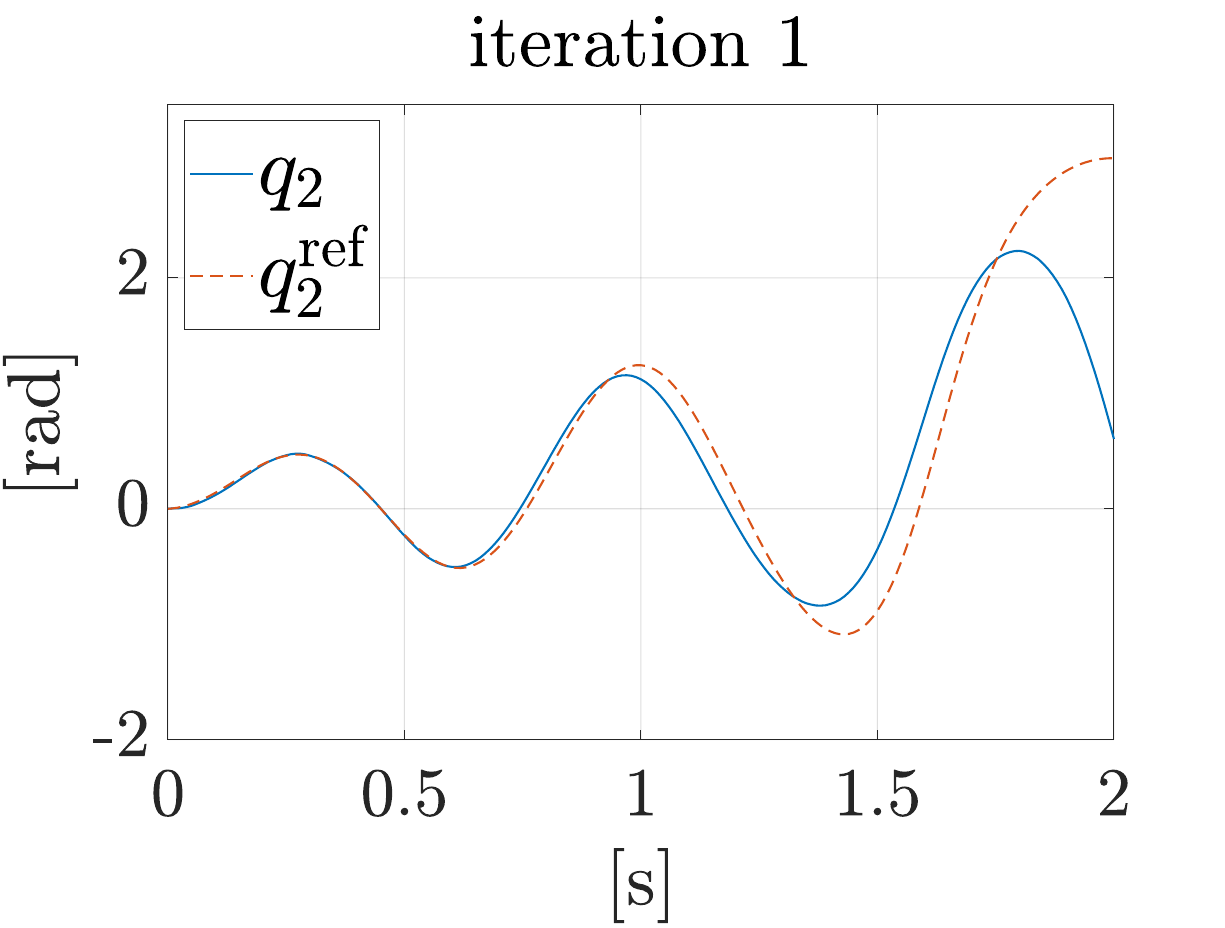}
\includegraphics[width=0.235\textwidth]{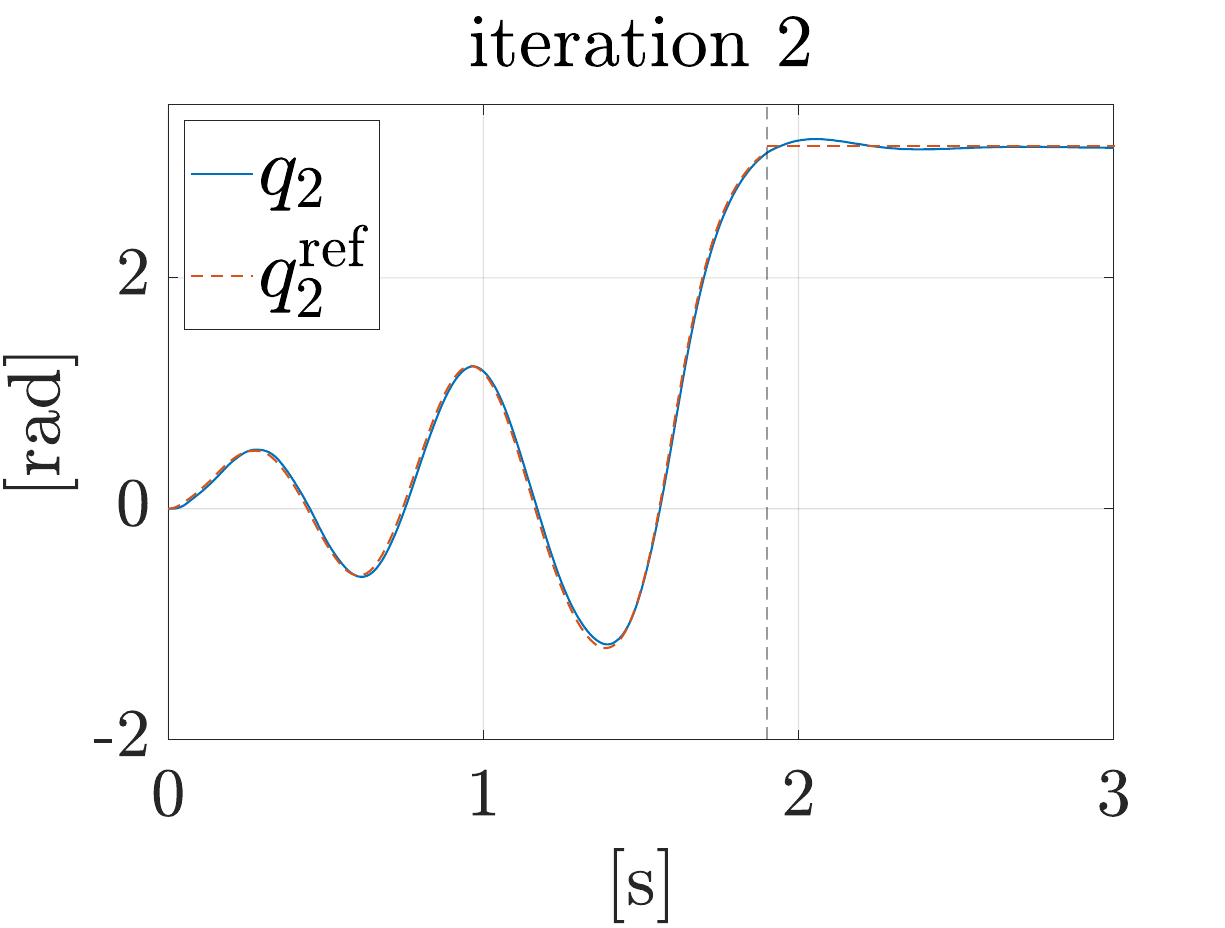}
}}

\usepackage{xcolor}
\definecolor{lgray}{gray}{0.30}
\usepackage{eso-pic}
\usepackage{hyperref}
\AddToShipoutPictureBG*{%
  \AtPageUpperLeft{%
    \setlength{\unitlength}{1mm}%
    \put(-18.5,-9){\makebox(\paperwidth,0)[c]{\parbox{0.8\textwidth}{IEEE Robotics and Automation Letters, 2022, vol. 7, no. 1, pp. 358-365}}}
  }
}

\AddToShipoutPictureBG*{%
  \AtPageUpperLeft{%
    \setlength{\unitlength}{1mm}%
    \put(-9.5,-14){\makebox(\paperwidth,0)[c]{\parbox{0.9\textwidth}{DOI: \href{https://ieeexplore.ieee.org/document/9610177}{10.1109/LRA.2021.3126899}}}}
  }
}

\title{
On-Line Learning for Planning and Control \\
of Underactuated Robots with Uncertain Dynamics}

\author{Giulio Turrisi, Marco Capotondi, Claudio Gaz, Valerio Modugno, Giuseppe Oriolo, Alessandro De Luca%
\thanks{Manuscript received July 9, 2021; Revised  August, 25, 2021; Accepted October 23, 2021.}
\thanks{This paper was recommended for publication by Associate Editor T. Asfour and Editor L. Pallottino upon evaluation of the reviewers' comments.
\emph{(Corresponding author: Marco Capotondi)}}
\thanks{The authors are with the Dipartimento di Ingegneria Informatica, Automatica e Gestionale, Sapienza Universit\`a di Roma, via Ariosto 25, 00185 Roma, Italy.  E-mail:
\{lastname\}@diag.uniroma1.it.}%
}

\begin{document}

\maketitle

\begin{abstract}
We present an iterative approach for planning and controlling  motions of underactuated robots with uncertain dynamics. At its core, there is a learning process which estimates the perturbations induced by the model uncertainty on the active and passive degrees of freedom. The generic iteration of the algorithm makes use of the learned data in both the planning phase, which is based on optimization, and the control phase, where partial feedback linearization of the active dofs is performed on the model updated on-line. The performance of the proposed approach is shown by comparative simulations and experiments on a Pendubot executing various types of swing-up maneuvers. Very few iterations are typically needed to generate dynamically feasible trajectories and the tracking control that guarantees their accurate execution, even in the presence of large model uncertainties.
\end{abstract}

\begin{IEEEkeywords}
Underactuated Robot, Model Learning for Control, Optimization and Optimal Control
\end{IEEEkeywords} 

\section{Introduction}
\label{sec:introduction}

\IEEEPARstart{U}{nderactuation} in mechanical systems occurs when there are less independent actuation inputs than generalized coordinates. This situation may be due to the nature of the mechanism, to its prevailing design, or it may be the result of an intentional choice aimed at reducing weight, cost or energy consumption. Many advanced robotic platforms are indeed underactuated, including manipulators with some passive joints, most underwater and aerial vehicles, legged robots, and nonprehensile manipulation systems.

An adverse effect of underactuation is that generic state space trajectories become unfeasible, since the dynamics of the passive degrees of freedom represents a set of second-order differential constraints that must be satisfied throughout any motion~\cite{Oriolo1991}; in practice, this limits the directions of instantaneous accelerations that can be commanded to the system. 
As a consequence, trajectory planning in the absence of obstacles, which is a relatively trivial issue for fully actuated robots, becomes a challenging problem in the presence of underactuation.

Motion control is also made more difficult by underactuation. One way to appreciate this is to consider that the dynamic models of fully actuated robots can always be made exactly linear and decoupled by using static feedback linearization~\cite{Isidori1995}, provided that an accurate model of the robot dynamics is available. In underactuated robots this cannot be achieved, and one has to deal directly with --- actually, make use of ---  nonlinear, coupled dynamic effects.

In the literature, several model-based techniques have been proposed for planning and stabilizing motions of specific underactuated robots, with notable emphasis on manipulators with passive joints~\cite{DeLuca2003}. In particular, the problem of state transfer between equilibria has been addressed mainly on two benchmark platforms, i.e., the Pendubot and the Acrobot; these are both 2R robots moving in the vertical plane with a single actuated joint (respectively, the first and the second). 
A classical approach is to use collocated or non-collocated 
Partial Feedback Linearization (PFL) in combination with energy-based controllers~\cite{Spong1994,Spong1998}. Swing-up maneuvers of these robots have been achieved using passivity-based approaches~\cite{fantoni02_chap,shiriaev02}, orbit stabilization~\cite{Orlov2006}, impulse-momentum techniques~\cite{Albahkali2009}, and sequential action control~\cite{Ansari2016}. Typically, the maneuver includes a final balancing phase realized through a Linear Quadratic Regulator (LQR) designed around the target equilibrium.

\begin{figure}[!t]
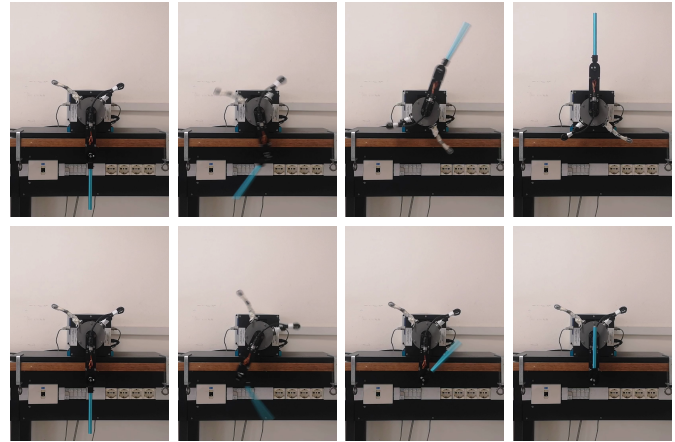

\SwingUpPend
\caption{The Pendubot performing two different swing-up maneuvers using the proposed method: the targets are the up-up equilibrium (first row) and the down-up equilibrium (second row). See the accompanying video.}
\label{fig:swing_up_pend}
\vspace{-15pt}
\end{figure}

Although effective, the above approaches have two main limitations from the viewpoint of this paper. First, the design techniques used for trajectory planning and  control are invariably specific (or had to be specialized) for the considered robot, and sometimes also for the particular maneuver. Second, and even more important, all of them require an accurate knowledge of the robot dynamic model for successful performance. Exceptions are~\cite{Qian2007,Moussaoui2016}, which propose robust control of the Pendubot via adaptive and fuzzy sliding modes respectively; however, these methods are tailored to the platform and do not tolerate large model uncertainties in practice.

To avoid the need for an accurate dynamic model, modern learning techniques have been applied for deriving feedforward and feedback control of robots~\cite{peters16}. 
In~\cite{nguyentuong2010}, a semi-parametric regression is used to reconstruct the inverse dynamics of a manipulator. 
In~\cite{lillicrap2016}, \cite{Deisenroth2011,Chatzilygeroudis2017} and \cite{saveriano2017}, Reinforcement Learning (RL) procedures are proposed to generate robot control policies in a data-efficient way. However, this class of algorithms is not able in general to ensure satisfaction of hard constraints in a specific robot task.
Along the same lines, the authors in~\cite{arcari2020} propose a meta-learning approach for a domain adaptation problem; in a new scenario, only the regressor weights are updated while the set of basis functions is kept the same, ultimately limiting the effectiveness of the model correction.
An optimization-based iterative learning approach is used in~\cite{schoellig2009, Mueller2012}, where experience from previous robot trials is used to build incrementally the feedforward command needed to follow a desired output trajectory.
In~\cite{capotondi2019}, we have presented a learning-based approach for trajectory tracking which relies on the existence of a nominal feedback linearizing control law for the robotic system. Similar techniques are proposed in~\cite{westenbroek20,greeff21}. These works, however, assume full actuation capability or at least feedback linearizability via dynamic feedback~\cite{murray95}. 

Other works that are more closely related to our approach have been published recently.
In~\cite{gillen20}, a method for the swing-up of an Acrobot has been proposed which avoids the need for a model by using deep RL, requiring however a huge number of experiments for training. 
A robust control scheme for trajectory tracking under repetitive disturbances has been presented in~\cite{zhang2019} for a 3R planar manipulator with two actuators and one passive joint. The control design is tailored to this specific system, and cannot be easily extended to generic underactuated robots. In~\cite{chen2019}, a learning scheme is proposed to realize trajectory tracking of underactuated balance robots (e.g., a Furuta pendulum); because of the simpler balancing task, the reference trajectory of the active joints is not replanned and stabilization in the large is never addressed.

In this paper, we build upon our learning method for fully actuated robots~\cite{capotondi2019} to devise an iterative approach for planning and controlling transfers between (stable or unstable) equilibria of underactuated robots in the presence of large dynamic uncertainties. The basic idea is to alternate off-line optimization-based planning and on-line PFL control, using regression to learn model corrections for the active and passive degrees of freedom.
As a result, dynamically feasible state reference trajectories are learned and convergence to zero trajectory tracking error is obtained over the iterations. 
The main benefits of the proposed approach are:
\begin{itemize}
    \item it applies to any underactuated robot;
    \item it applies to any state transfer maneuver;
    \item convergence is reached even in the presence of large uncertainties on the robot dynamics, requiring very few iterations in the considered case studies; 
    \item more accuracy in the nominal dynamic model leads to even faster convergence;
    \item additional constraints (on state, on input, obstacle avoidance, etc.) can be explicitly taken into account in the optimization problem of the planning phase.
\end{itemize}

As an application, we provide an extensive evaluation of the performance of our approach on a Pendubot which must execute various swing-up maneuvers and state transfers between unstable equilibria (see Fig.~\ref{fig:swing_up_pend}).

The paper is organized as follows. Section~\ref{sec:problem_formulation} introduces the dynamic model of underactuated robots, highlighting how model uncertainties affect the active and passive subsystems. The proposed iterative approach is presented in Sec.~\ref{sec:iterative_planning_and_control}, discussing both the planning and the control phases and describing the data collection procedures and the regressors adopted for learning. In Sec.~\ref{sec:results}, we report on the application to the Pendubot, showing comparative simulation and experimental results. Finally, some general conclusions about the approach are drawn in Sec.~\ref{sec:conclusions}.

\section{Problem Formulation}
\label{sec:problem_formulation}

For a robot with $n$ degrees of freedom (dof) and $m < n$ actuators, the dynamics can be expressed~\cite{DeLuca2003} as
\begin{eqnarray} 
\Mm_{aa}(\qv) \ddqv_{a} +  \Mm_{ap}(\qv) \ddqv_{p}+ \nv_{a}(\qv,\dqv) &\!\!\!\!=\!\!\!\!& \tauv \label{eq:dyn_model_underactuateda}\\
\Mm_{pa}(\qv) \ddqv_{a} +  \Mm_{pp}(\qv) \ddqv_{p}+ \nv_{p}(\qv,\dqv) &\!\!\!\!=\!\!\!\!& \zerov,
\label{eq:dyn_model_underactuatedu}
\end{eqnarray}
where $\qv = (\qv_{a},\qv_{p})$ is the $n$-dimensional configuration vector, with $\qv_{a}$, $\qv_{p}$ representing respectively the $m$ active and the $n-m$ passive generalized coordinates. The inertia matrix $\Mm$ and the vector $\nv$ of the remaining nonlinear terms are partitioned accordingly. The $m$ generalized forces $\tauv$ only perform work on the $\qv_a$ coordinates. We do not assume any structural control property (e.g., feedback linearizability or flatness) for system~(\ref{eq:dyn_model_underactuateda}-\ref{eq:dyn_model_underactuatedu}), nor any particular degree of underactuation.

In the presence of model perturbations (incorrect parameters and/or unmodeled dynamics), we can write
\begin{equation}
\Mv = \hat{\Mv} + \Delta \Mv \qquad
\nv = \hat{\nv} + \Delta \nv.
\label{eq:perturbations}
\end{equation}
Only the nominal terms $\hat{\Mv}$ and $\hat{\nv}$ are known and available for control design.

Let $\xv=(\qv,\dot \qv)$ be the robot state. 
Given a start and a goal equilibrium points, respectively denoted by $\xv_s=(\qv_s,\zerov)$ and $\xv_g=(\qv_g,\zerov)$, we want to plan and execute in a fixed time $T$ a transfer motion from the start to the goal, while satisfying constraints on state and/or inputs, collectively expressed in the form $\hv (\qv, \tauv) \leq \zerov$. 
This {\em transfer between equilibria} problem is particularly challenging for robots that are not fully actuated because not all trajectories between two equilibria are feasible.

For the following developments, it is convenient to perform a preliminary nonlinear feedback aimed at exactly linearizing the {\em nominal} active dynamics. 
This collocated PFL controller is always well-defined and takes the form
\begin{equation}
\tauv_{\textrm{PFL}} = \left(\hat\Mm_{aa} -\hat\Mm_{ap}\hat\Mm^{-1}_{pp}\hat\Mm_{pa} \right) \uv + \hat\nv_{a} -\hat\Mm_{ap}\hat\Mm_{pp}^{-1}\hat\nv_{p},
\label{eq:tau_pfl}
\end{equation}
where $\uv \in \Real^m$ is the new input, i.e., the acceleration of the active dofs.

Using~(\ref{eq:tau_pfl}) and~(\ref{eq:perturbations}) in~(\ref{eq:dyn_model_underactuateda}--\ref{eq:dyn_model_underactuatedu}), we obtain the perturbed closed-loop dynamics
\begin{eqnarray}
\ddqv_{a} 
&\!\!\!\!=\!\!\!\!& \uv + \deltav_{a}(\qv, \dqv,\uv) 
\label{eq:model_uncertanty_delta_a}
\\
\ddqv_{p} &\!\!\!\!=\!\!\!\!& -\hat{\Mm}^{-1}_{pp} ( \hat{\nv}_{p} + \hat{\Mm}_{pa} \ddqv_{a}) + \deltav_{p}(\qv, \dqv,\ddqv_{a}),
\label{eq:model_uncertanty_delta_u}
\end{eqnarray}
where $\deltav_{a}$ and $\deltav_{p}$ represent the cumulative effect of perturbations on the active and passive subsystems, respectively. 

\section{The Proposed Iterative Approach}
\label{sec:iterative_planning_and_control}

\begin{figure*}[t]
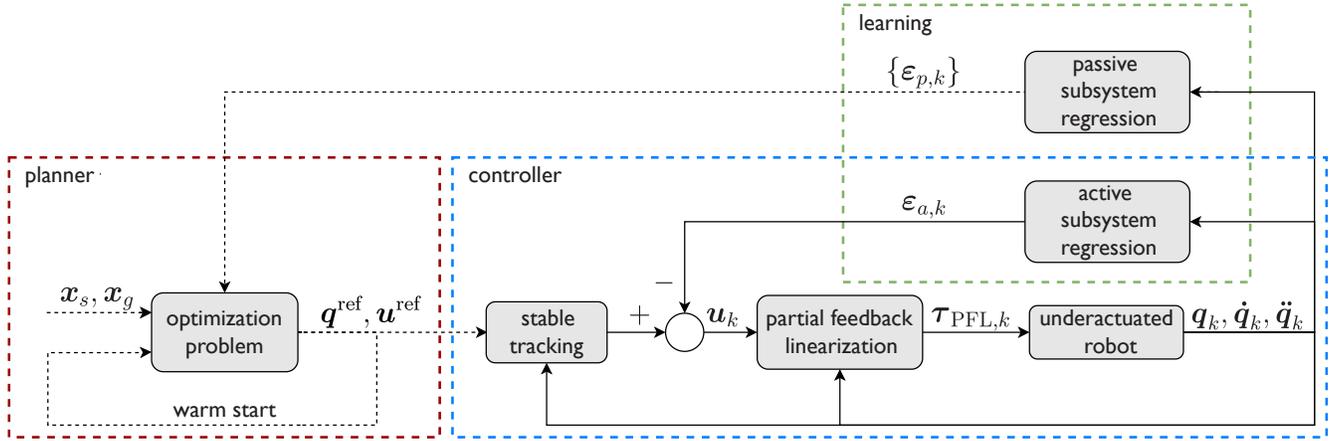

\BlockDiagram
\caption{Block diagram of the generic iteration of the proposed algorithm. Solid signal lines represent data that are used at each time step, whereas dashed lines are data transferred at the end of the iteration.}
\label{fig:block_diagram}
\end{figure*}

The presence of model perturbations affects the considered planning and control problem at two levels. First, planning based on the nominal model would produce trajectories that may not be feasible, and in any case do not land at the goal equilibrium. Second, even when the reference trajectory is feasible, effective tracking is not achieved if the controller is designed on the nominal model.   

In this section, we describe an iterative scheme for concurrent planning and control. At its core there is a {\em learning process} (Sects.~\ref{sec:regressor_modeling}--\ref{sec:learning_passive_joint}) which continuously updates two regressors $\epsilonv_{a}$ and $\epsilonv_{p}$, respectively estimates of the perturbations $\deltav_{a}$ and $\deltav_{p}$ in~(\ref{eq:model_uncertanty_delta_a}--\ref{eq:model_uncertanty_delta_u}). Both regressors are reconstructed from position measurements during robot motion. 

Each iteration consists of an off-line \emph{planning} phase and~an on-line \emph{control} phase. In the planning phase (Sect.~\ref{sec:planning}), the nominal model is corrected by taking into account $\epsilonv_{p}$; an optimization problem is then solved to compute a reference trajectory $\qv^{\rm ref}(t)$ leading this model to $\xv_g$ at time $T$.
In the control phase (Sect.~\ref{sec:control}), the robot tracks $\qv^{\rm ref}(t)$ under the action of a PFL control law given by~(\ref{eq:tau_pfl}), in which the corrective term $\epsilonv_{a}$ is added to the commanded acceleration $\uv$.
During the motion, new data points are collected and used in the learning process. 

A block diagram of the generic iteration of the proposed approach is shown in Fig.~\ref{fig:block_diagram}.

\subsection{Planning}
\label{sec:planning}

In the planning phase, a reference trajectory is computed by solving a numerical optimal control problem for the underactuated robot. In particular, a prediction model is obtained by setting $\deltav_a=\zerov$ and $\deltav_p=\epsilonv_p$ in eqs.~(\ref{eq:model_uncertanty_delta_a}--\ref{eq:model_uncertanty_delta_u}):
\begin{eqnarray}
\ddqv_{a} &\!\!\!\!=\!\!\!\!& \uv \label{eq:model_planning_control_a}\\
\ddqv_{p} &\!\!\!\!=\!\!\!\!& -\hat{\Mm}^{-1}_{pp} ( \hat{\nv}_{p} + \hat{\Mm}_{pa} \uv) + \epsilonv_{p}(\qv, \dqv,\uv).
\label{eq:model_planning_control_u}
\end{eqnarray}
In other words, we are assuming in~(\ref{eq:model_planning_control_a}) that partial feedback linearization has been achieved in spite of model perturbations. The rationale is that the control law will try to cancel $\deltav_a$ as much as possible using a correction term equal to its current estimate $\epsilonv_{a}$ (see Sect.~\ref{sec:control}). Moreover, the available estimate $\epsilonv_{p}$ of the perturbation on the passive subsystem has been used in~(\ref{eq:model_planning_control_u}). 
Upon convergence of the overall scheme, eq.~(\ref{eq:model_planning_control_a}) will become exact, and $\epsilonv_{p}$ in~(\ref{eq:model_planning_control_u}) will eventually be equal to $\deltav_p$.

In principle, we could have also included $\epsilonv_{a}$ in the right-hand side of~(\ref{eq:model_planning_control_a}). The learning transient would be similar and, upon convergence, the obtained system behavior would be the same. However, the separate use of one regressor ($\epsilonv_{p}$) in the planning phase and of the other ($\epsilonv_{a}$) in the control phase proves to be computationally more efficient.

We consider a discrete-time setting in which the input $\uv$ is piecewise-constant over $N$ sampling intervals of duration $T_s=T/N$. Denoting by $\fv(\cdot)$ a discretization of the state-space representation corresponding to~(\ref{eq:model_planning_control_a}--\ref{eq:model_planning_control_u}), with the robot state $\xv_i=\xv(t_i)$ and the starting and goal equilibrium points $\xv_{s}$ and $\xv_{g}$ defined in Sec.~\ref{sec:problem_formulation}, the optimization problem (OP) is written as
\[
\min_{\uv_0,\dots,\uv_{N-1}} \sum^{N-1}_{i=0} J(\xv_i,\uv_i) +  J_N(\xv_N)
\]
subject to
\begin{eqnarray*}
\xv_{i+1}-\fv(\xv_i,\uv_{i}) &\!\!\!=\!\!\!& 0, \qquad i=0 \dots,N-1, \\
\gv(\xv_i) &\!\!\!\leq\!\!\!& 0, \qquad i=1,\dots,N,\\
\hv(\uv_i) &\!\!\!\leq\!\!\!& 0, \qquad i=1,\dots,N-1,
\end{eqnarray*}
with $\xv_0 = \xv_s$.
The objective function is the sum of a stage cost $J$ and a terminal cost $J_N$, both penalizing the state error with respect to the goal $\xv_g$ and the control effort, while $\gv$ and $\hv$ represent state and input constraints, respectively. 
The cost terms take the form
\begin{eqnarray*}
J(\xv_i,\uv_i) &\!\!\!=\!\!\!& \|\xv_g-\xv_i\|^{2}_{\Qm} \ + \|\uv_i\|^{2}_{\Rm},
\\[2pt]
J_N(\xv_N) &\!\!\!=\!\!\!& \|\xv_g-\xv_N\|^2_{\Qm_N}.
\end{eqnarray*}
where $\Qm$, $\Qm_N$ and $\Rm$ are positive-definite, symmetric matrices of weights.
The solution of OP is a reference trajectory with the associated nominal input, represented by discrete sequences $\qv^{\rm ref}=\{\qv^{\rm ref}_1,\dots,\qv^{\rm ref}_N\}$ and $\uv^{\rm ref}=\{\uv^{\rm ref}_0,\dots,\uv^{\rm ref}_{N-1}\}$ respectively. The reference velocities $\dot \qv^{\rm ref}=\{\dot \qv^{\rm ref}_1,\dots,\dot \qv^{\rm ref}_N\}$ are also available. 

To speed up convergence to a solution, one typically uses the reference trajectory of the previous iteration as a warm start when solving the current OP.

\subsection{Control} 
\label{sec:control}

 In the control phase, the robot moves under the action of a digital\footnote{For simplicity, it is assumed that the control sampling interval is the same $T_s$  used for planning.} control law aimed at driving $\qv$ along the current reference trajectory $\qv^{\rm ref}$. To achieve stable tracking of $\qv^{\rm ref}_a$, the commanded acceleration $\uv_k$ in $[t_k,t_{k+1})$ is chosen as
\begin{equation}
\uv_k = \uv^{\rm ref}_k + \Km_P (\qv^{\rm ref}_{a,k} - \qv_{a,k}) + \Km_D (\dot \qv^{\rm ref}_{a,k} - \dot \qv_{a,k}) - \epsilonv_{a,k},
\label{eq:learned_acceleration}
\end{equation}
with $\Km_P,\Km_D>0$. Here, the nominal input produced by the planner is used as feedforward term, and the current regressor $\epsilonv_{a,k}$ has been added to cancel at best the perturbation $\deltav_a$ affecting the active subsystem~(\ref{eq:model_uncertanty_delta_a}). Note that as soon as $\qv_a$ will be able to follow exactly $\qv^{\rm ref}_a$, the passive variables $\qv_p$ will evolve as planned in the previous phase. 

Next, we use~(\ref{eq:tau_pfl}) to compute the generalized force as 
\begin{equation}
\tauv_{\textrm{PFL},k} = \hat \Bm_k \uv_k + \hat \etav_k,
\label{eq:learned_PFL}
\end{equation}
where 
\[
\hat \Bm_k =\hat\Mm_{aa}(\qv_k) -\hat\Mm_{ap}(\qv_k)\hat\Mm^{-1}_{pp}(\qv_k)\hat\Mm_{pa}(\qv_k)
\]
and
\[
\hat \etav_k = \hat\nv_{a}(\qv_k,\dot \qv_k) -\hat\Mm_{ap}(\qv_k)\hat\Mm_{pp}^{-1}(\qv_k)\hat\nv_{p}(\qv_k,\dot \qv_k).
\]

\subsection{Regressors for estimating perturbations}
\label{sec:regressor_modeling}

In this work, we employ Gaussian Processes (GP) regressors \cite{rasmussen2006} for reconstructing $\epsilonv_{a}$ and $\epsilonv_{p}$, given the good performance that this technique displays in the on-line learning context. However, it is important to notice that other techniques such as Neural Networks, Generalized Linear Regression or Support Vector Machine could have been adopted without any modifications on the structure of the framework. 

Considering $\mathcal{D}=\left\{\left(\Xm_i, Y_{i}=\phi\left(\Xm_{i}\right)+\omegav_{i}\right) | 1 \leq i \leq n_d\right\}$ representing a set of input-output noisy observations where $\omegav \sim \mathcal{N}(\zerov,\,\Sigmam_{\omega})$ and $\phi$ indicates a unidimensional function to reconstruct.
With $\Km(\Xm,\Xm)$ we define the covariance matrix whose elements are computed through the inner product operation $k(\cdot,\cdot)$ and $\kv^{T}(\cdot)$ is a row vector obtained by computing $k$ between a new point and every element of $\Xm$. Let us denote the value of the function to reconstruct at an arbitrary point as $Y_{n_d+1} = \phi(\Xm_{n_d+1})$. Exploiting the properties of  Gaussian processes, the ensemble of observation in the dataset defined as $\Ym_{1:n_d}$ and $Y_{n_d+1}$ are jointly Gaussian:
\begin{equation*}
    \left[\begin{array}{c}
{\Ym}_{1: n_d} \\
Y_{n_d+1}
\end{array}\right] \sim \mathcal{N}\left(\mathbf{0},\left[\begin{array}{cc}
\Km & \kv \\
\kv^{T} & k\left(\Xm_{n_d+1}, \Xm_{n_d+1}\right)
\end{array}\right]\right)
\end{equation*}
It is possible to define the predictive distribution that approximates the perturbation $\delta(\hat{\Xm})$ for a generic query point $\hat{\Xm}$ as
\begin{equation*}
    \varepsilon(\hat{\Xm} | \mathcal{D}) \sim \mathcal{N}\left({\mu}(\hat{\Xm}),{\sigma}^{2}(\hat{\Xm})\right)
\end{equation*}
where
\begin{align*}
{\mu}(\hat{\Xm}) &= \kv^{T\!}(\hat{\Xm})(\Km+\Sigmam_{\omega})^{-1} \Ym \\
{\sigma}^{2}(\hat{\Xm}) & = k(\hat{\Xm},\hat{\Xm}) -\kv^{T\!}(\hat{\Xm})(\Km+\Sigmam_{\omega})^{-1}\kv(\hat{\Xm}).
\end{align*}
In this context, $\mu$ represents the regressor prediction while ${\sigma}^{2}(\cdot)$ describes the epistemic error associated to $\mu(\cdot)$. In this work, for reconstructing a multidimensional observation, we stack a set of unidimensional GPs.

Since no assumption is made on the structure of the unmodeled dynamics, we employ as $k(\cdot,\cdot)$ a squared exponential kernel defined as
\[
k\left(\Xv_{i}, \Xv_{j}\right)=a^{2}\exp \left(- \frac{\left\|\Xv_{i}-\Xv_{j}\right\|^{2}}{2\,l^{2}}\right),
\]
where the length-scale $l$ and the amplitude $a$ represent the hyperparameters of the GP regressor.

\subsection{Dataset collection procedure for the active dofs}
\label{sec:learning_actuated_joint}

We now show how to collect the data points that are used to learn the estimate $\epsilonv_{a,k}$ to be used at each instant in the commanded acceleration~(\ref{eq:learned_acceleration}). 

The idea is to perform a regression based on the difference between the commanded acceleration and the actual acceleration for the actuated dofs. In fact, from eq.~(\ref{eq:model_uncertanty_delta_a}) we may write
\begin{equation}
\deltav_{a,k} = \ddot \qv_{a,k} - \uv_{k}.
\label{eq:delta_ak}
\end{equation}
In view of eq.~(\ref{eq:delta_ak}), a new data point is generated at the $k$-th control step as
\[
\Xv_{a,k} = (\qv_{k}, \dqv_{k}, \uv_k)  \qquad
\Yv_{a,k} = \ddot \qv_{a,k} - \uv_{k}.
\]
with the acceleration $\ddot \qv_{a,k}$ to be reconstructed numerically. We note that the actual acceleration is functionally dependent through~(\ref{eq:model_uncertanty_delta_a}) on the robot state $(\qv_{k}, \dqv_{k})$ and on the commanded acceleration $\uv_k$, i.e., on the input $\Xv_{a,k}$ of the regression scheme. 

Every time a new data point is available, it is immediately used to update the regressor $\epsilonv_{a}$. However, the hyperparameters of the kernel function are only updated at the end of each iteration.

A potential issue of GP regression is that the computational complexity of the prediction is $\mathcal{O}(n_{d}^{3})$, with $n_{d}$ the size of the dataset. 
To keep the computation of ${\epsilonv_{a}}$ fast enough for real-time control, an approximate regression is performed using a reduced set of only $d$ datapoints, chosen on the basis of the information gain criterion \cite{seeger2003}. This leads to a reduced complexity $\mathcal{O}(d^{2} \cdot n_{d})$.

\subsection{Dataset collection procedure for the passive dofs}
\label{sec:learning_passive_joint}

To learn an estimate $\epsilonv_p$ of the model perturbation $\deltav_{p}$, we compare the commanded and the actual acceleration for the passive dofs.

In fact, from  eq.~(\ref{eq:model_uncertanty_delta_u}) we have
\begin{equation}
\deltav_{p,k} = \ddqv_{p,k} + \hat{\Mm}^{-1}_{pp,k} ( \hat{\nv}_{p,k} + \hat{\Mm}_{pa,k} \,\ddqv_{a,k}).
\label{eq:delta_pk}
\end{equation}
Given numerical approximations of the actual accelerations $\ddqv_{a,k}$ and $\ddqv_{p,k}$, a new data point is generated at the $k$-th step as 
\[
\begin{array}{rcl}
\Xv_{p,k} &\!\!\!\!=\!\!\!\!& (\qv_{k}, \dqv_{k}, \ddqv_{a,k}) 
\\[3pt]
\Yv_{p,k} &\!\!\!\!=\!\!\!\!& \ddqv_{p,k} + \hat{\Mm}^{-1}_{pp,k} ( \hat{\nv}_{p,k} + \hat{\Mm}_{pa,k} \,\ddqv_{a,k}).
\end{array}
\]

Differently from $\epsilonv_{a}$, {\em all} the data points computed during the iteration are used to update the passive subsystem regressor $\epsilonv_{p}$ at the end of each trial; in fact, since the planning phase is performed off-line, the complexity associated to exact regression does not represent a problem here. As before, the hyperparameters of the kernel function are also updated at the end of the iteration.

\section{Application to the Pendubot}
\label{sec:results}

\begin{figure}[t]
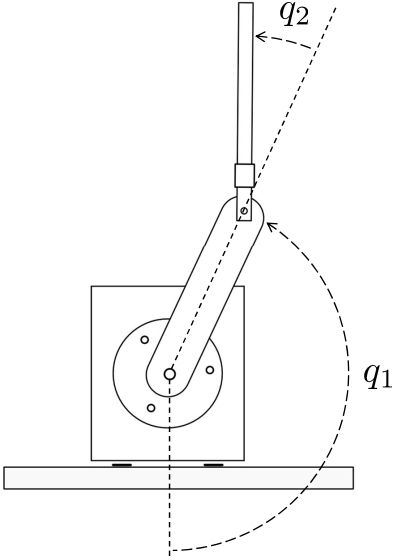

\Pendubot
\vspace{-5pt}
\caption{The Pendubot and its generalized coordinates.}
\label{fig:Pendubot}
\vspace{-10pt}
\end{figure}

The proposed approach has been validated through simulations and experiments on the Pendubot, a  two-link arm moving in the vertical plane with an active joint at the shoulder and a passive joint at the elbow ($\qv_a=q_1$ and $\qv_p=q_2$). See Fig.~\ref{fig:Pendubot} for the definition of the generalized coordinates and~\cite{Turrisi20} for the dynamic model of the Pendubot, complete with nominal parameter values for our prototype.

In the following, we will address the problem of executing various transfer motions between equilibria in the presence of severe uncertainty on the dynamic model. The proposed iterative method is used to steer the Pendubot to the basin of attraction of an LQR balancing controller designed around the goal state. The latter is obviously needed to stabilize the robot after the planning horizon $T$. 

The discretized state-space model used in the planning phase has been obtained by Euler method. The sampling interval is set to $T_s=10$~ms. The cost function $J$ in OP includes two quadratic terms that penalize the state error with respect to the goal $\xv_g=(q_{1,g},q_{2,g},0,0)$ as well as the control effort. Optimization is performed in MATLAB using the {\tt fmincon}  function, which implements a Sequential Quadratic Programming method. The joint velocities are bounded as $|\dot q_1| \le 8$ rad/s and $|\dot q_2| \le 15$ rad/s. Finally, terminal constraints are included to guarantee convergence at time $T$ to the following basin of attraction of the balancing controller
\[
\left\vert q_{j,N}- q_{j,g}\right\vert \le 0.2, \qquad
\left\vert \dot q_{j,N} \right\vert \le 0.5, \qquad j=1,2,
\]
which was found to be adequate for all goal states.

In the control phase, the PD gains in~(\ref{eq:learned_acceleration}) are chosen as $K_P=50$ and $K_D=20$, while the sampling interval is again 10~ms. 

While all data points (with $n_d$ equal to $N$ times the number of iterations so far) are considered for updating $\epsilonv_p$, the maximum number of data points used for computing $\epsilonv_a$ in real time is $d=180$.

Refer to the accompanying video for clips from all simulations and experiments shown in the following.

\subsection{Simulation results}
\label{sec:results_simulation}

Two scenarios of transfer between equilibrium states will be presented. To show that the proposed method can achieve robust performance in the presence of severe model perturbations, we perturbed for control design the nominal values in~\cite{Turrisi20}, increasing by $30\%$ the link masses $m_{1}$ and $m_{2}$ and reducing by the same percentage the distances $a_1$ and $a_2$ of the centers of mass of the two links from their respective joints. The link barycentral inertias $I_1$ and $I_2$ were changed accordingly.

\begin{figure}[t]
\SwingUpSim
\caption{Simulation scenario 1 ({\em swing-up to} $\qv^{\,\text{u-u}}$): results without learning. Left: Using the nominal model for planning and the true model for control. Right: Vice versa.}%
\label{fig:swingup_sim}%

\bigskip

\IterationsSim
\caption{Simulation scenario 1 ({\em swing-up to} $\qv^{\,\text{u-u}}$): results with the proposed approach. Just before the end of the third iteration, the state has converged to the basin of attraction of the balancing controller, which is then activated (as indicated by the vertical dashed line).} 
\label{fig:iterations_sim}

\medskip

\ShiriaevSim
\caption{Simulation scenario 1 ({\em swing-up to} $\qv^{\,\text{u-u}}$): results with the method in~\cite{shiriaev02}. Left: assuming exact model knowledge the state enters the basin of attraction of the LQR controller at $t=1.3$~s circa. Right: with the same model uncertainty of Fig.~\ref{fig:iterations_sim}, convergence is not achieved.}.
\label{fig:shiriaev_sim}
\vspace{-15pt}
\end{figure}

In the first scenario, the start configuration is $\qv_s = (0,0)$ while the goal is the \emph{up-up} configuration $\qv_g=\qv^{\,\text{u-u}} = (\pi, 0)$, corresponding to a transfer from a stable to an unstable equilibrium ({\em swing-up}). The planning horizon is chosen as $T=1.6$~s ($N=160$). 

To highlight the necessity of learning in both the planning and control phases, we have preliminarily considered two complementary situations where learning is not used.  Figure~\ref{fig:swingup_sim}, left, refers to the first situation, in which we use the nominal model for planning and the true model for control. The result shows that planning the motion of an underactuated robot based on an inaccurate model produces dynamically unfeasible trajectories, that cannot be tracked in spite of the ideality of the controller. Vice versa, in Fig.~\ref{fig:swingup_sim}, right, the true model is used for planning and the nominal for control. As expected, the inaccuracy of the controller prevents the completion of the swing-up maneuver.

Next, we tested the proposed approach on the same scenario, obtaining the results in Fig.~\ref{fig:iterations_sim}. 
After three iterations, the Pendubot is able to track with sufficient accuracy the planned trajectory, ultimately entering the basin of attraction of the balancing controller to complete the swing-up maneuver. This shows that, in spite of the very large model uncertainty, the learning component of our method is able to reconstruct the correct model in  few iterations. Further iterations of the planning-control sequence do not change significantly the resulting motion.

To put our result in perspective, we have applied to this scenario also the passivity-based swing-up method proposed in~\cite{shiriaev02}, using the same balancing controller in the final phase. As shown in Fig.~\ref{fig:shiriaev_sim}, the method works perfectly if the robot model is exactly known, but is unable to complete the maneuver in the presence of the considered model uncertainty. In particular, while the first joint still converges to its target, the passive joint drifts away very quickly. 

\begin{figure}[t]
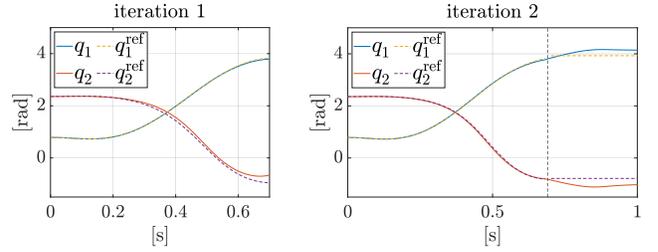

\IterationsSimForcedEq
\caption{Simulation scenario 2 (\emph{transfer between unstable equilibria}): results with the proposed approach. Two iterations are needed to achieve convergence.}
\label{fig:iterations_sim_forcedEq}
\vspace{-15pt}
\end{figure}

In the second scenario, the start is $\qv_s = (\pi/4,3 \pi/4)$ while the goal is $\qv_g = (5\pi/4,-\pi/4)$; this amounts to a transfer between two unstable equilibria. The planning horizon is chosen as $T=0.7$~s ($N=70$).
The results are shown in Fig.~\ref{fig:iterations_sim_forcedEq}. 
Two iterations are now sufficient to reach the basin of attraction of the balancing controller, thus completing the maneuver correctly. Indeed, a closer look at the joint motion (see also the accompanying video) reveals that in both iterations the transfer is performed with the second link approximately vertical, a situation which inherently reduces the effect of the uncertain dynamic parameters, leading to a faster convergence.

We have performed further simulations on a 3R Pendubot with {\em two} passive joints that must execute a swing-up maneuver to $\qv^{\,\text{u-u-u}}$ under perturbed conditions similar to scenario~1. Once again, convergence was achieved in 3 iterations, a result suggesting that our method performs effectively also for higher degrees of underactuation. See the accompanying video for an animated clip.

\subsection{Experimental results}
\label{sec:results_real}

The proposed method has also been tested experimentally on our Pendubot prototype, using again the nominal model in~\cite{Turrisi20} for planning and control design.
Joint velocities and accelerations are obtained in real time via filtered numerical differentiation of encoder measurements. To further remove the noise affecting the learning dataset for the passive dofs, we used a non-causal Savitzky-Golay filter to compute $\ddot q_2$. 

\begin{figure*}[t]
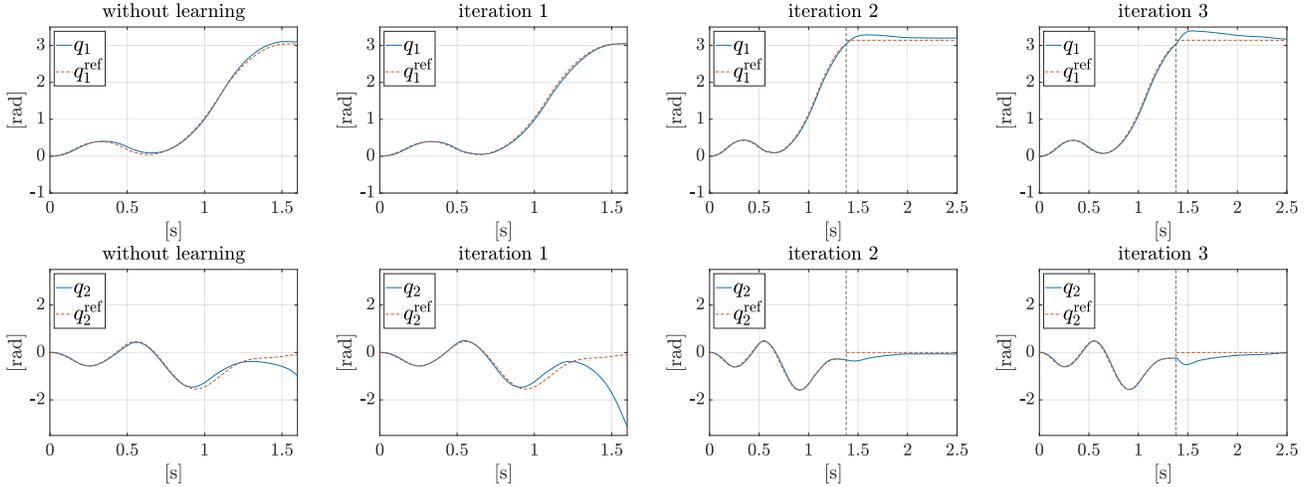

\ExperimentUpUp
\caption{Experimental scenario 1 ({\em swing-up to} $\qv^{\,\text{u-u}}$): results with the proposed approach. For comparison, the first column shows the results without learning, i.e., when partial feedback linearization and stable tracking for the first joint are computed on the nominal model.}
\label{fig:experiment_up_up}
\vspace{-12pt}
\end{figure*}

The first experiment replicates the swing-up scenario to $\qv^{\,\text{u-u}}$ of Sec.~\ref{sec:results_simulation}, using the same planning horizon of $T=1.6$~s ($N=160$). The results are shown in Fig.~\ref{fig:experiment_up_up}. 
Only two iterations are required for our method to enter the basin of attraction of the LQR controller.

\begin{figure}[t]
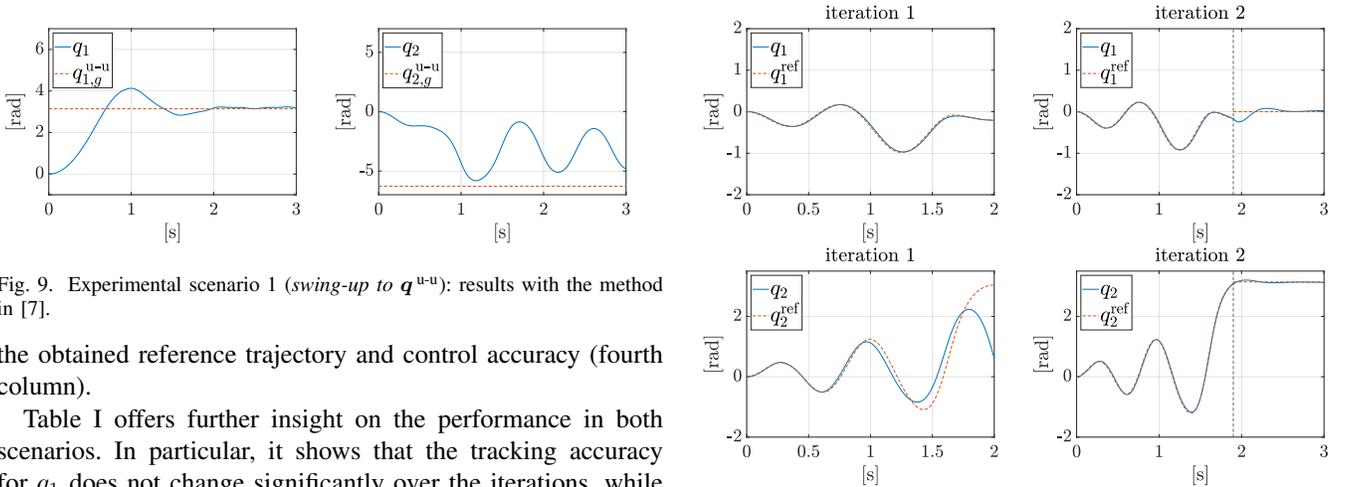

\ShiriaevExperiment
\caption{Experimental scenario 1 ({\em swing-up to} $\qv^{\,\text{u-u}}$): results with the method in~\cite{shiriaev02}.}
\label{fig:shiriaev_real}
\vspace{-15pt}
\end{figure}
 
The combination of on-line learning of the active joint dynamics together with the off-line re-planning of both joint trajectories, driven by the regressor built for the passive joint dynamics, allows a successful execution of the swing-up maneuver. When comparing the tracking errors between the single run without learning (Fig.~\ref{fig:experiment_up_up}, first column) and the first iteration of the method (Fig.~\ref{fig:experiment_up_up}, second column), no major changes are observed for the active joint $q_1$, whereas the passive joint $q_2$ behaves quite differently toward the end of the motion. In both cases, the error diverges (the second link falls in two opposite directions) because the currently planned trajectory is still dynamically unfeasible for the Pendubot. 
Already at the second iteration (third column), the robot is correctly driven to the basin of attraction of the desired equilibrium (the LQR stabilizer is triggered at about $t=1.4$~s).
Performing a third planning-control iteration shows no significant variation of the obtained reference trajectory and control accuracy (fourth column).

Table~\ref{tab:RMSE} offers further insight on the performance in both scenarios. In particular, it shows that the tracking accuracy for $q_1$ does not change significantly over the iterations, while the evolution of $q_2$ gets increasingly closer to the planned trajectory, as the latter approaches feasibility thanks to the model learning procedure. 

\begin{table}[b]
\centering
\caption{tracking rmse {\rm [rad]} in the experiments}
\label{tab:RMSE}
\begin{tabular}{*5c}
\toprule
{} &  \multicolumn{2}{c}{scenario 1} & \multicolumn{2}{c}{scenario 2}\\
\midrule
{}             & $q_1$   & $q_2$   & $q_1$ & $q_2$ \\[3pt] 
\cline{2-5} 
\\[-2pt]
without learning &  0.045  & 0.191   & 0.056  & 0.320  \\
iteration 1      &  0.035  & 0.623   & 0.022  & 0.470  \\
iteration 2      &  0.037  & 0.038   & 0.023  & 0.034  \\
iteration 3      &  0.036  & 0.036   &   --   &   --   \\
\bottomrule
\end{tabular}
\end{table}

For comparison, Fig.~\ref{fig:shiriaev_real} shows the experimental results obtained in this scenario with the method of~\cite{shiriaev02} under the same nominal information on the robot dynamic model. While the active joint converges to its desired goal, the second joint oscillates (with a light damping due to friction) without ever entering the basin of attraction of the stabilizing controller. Therefore, we can claim that the learning procedure makes the proposed method able to withstand a level of model uncertainty which is not tolerated by purely model-based controllers.

The second experiment is again a swing-up scenario, but the goal is now the \emph{down-up} configuration $\qv^{\,\text{d-u}} = (0, \pi)$. The planning horizon has been set to $T=2$~s ($N=200$). As shown in Fig.~\ref{fig:experiment_down_up}, also in this case the learning procedure allows to complete the maneuver successfully after two iterations (see also the second column in Table~\ref{tab:RMSE}).

\begin{figure}[t]
\ExperimentDownUp
\caption{Experimental scenario 2 ({\em swing-up to} $\qv^{\,\text{d-u}}$): results with the proposed approach. Just before the end of the second iteration the state converges to a region where the LQR controller can be successfully activated.}
\label{fig:experiment_down_up}
\vspace{-10pt}
\end{figure}

\section{Conclusions}
\label{sec:conclusions}

We have proposed an iterative method for planning and controlling motions of underactuated robots in the presence of model uncertainty. The method hinges upon a learning process which estimates the induced perturbations on the dynamics of the active and passive dofs. Each iteration includes an off-line planning phase and an on-line planning phase, which take advantage of the learned data to improve the feasibility of the planned trajectory and the accuracy of its tracking. 

The proposed approach was validated by application to the Pendubot, a well-known underactuated platform consisting of a 2R planar robot with a passive elbow joint. In particular, numerical simulations of our iterative method starting with considerable errors in the nominal dynamic parameters ($\pm$30\% of the true values) have shown that swing-up maneuvers and transfers between unstable equilibria can be executed successfully after very few iterations. This remarkable performance was confirmed in experimental tests on a real Pendubot. 

In addition to applicability to general underactuated systems and independence from the specific maneuver, a further aspect of our method that deserves to be emphasized is that no torque measurement is required. In fact, only positions, velocities and accelerations must be available, so that implementation is possible using only encoders. Another interesting feature is the possibility to incorporate constraints on the robot states and/or inputs in the planning phase, as well as to handle (without any modification) also the presence of repetitive external disturbances.

Future work will consider the problem of guaranteeing hard constraints during the entire learning transient, by explicitly taking into account the covariance of the uncertainty during the planning phase. Moreover, we plan to test the method on different underactuated robots, namely quadrotor UAVs and humanoids.

\balance 

\bibliographystyle{IEEEtran}
\bibliography{bibliography}

\end{document}